\documentclass[11pt]{article}

\PassOptionsToPackage{hyphens}{url}
\usepackage[preprint]{acl}

\usepackage{times}
\usepackage{latexsym}
\usepackage{amsmath,amssymb}
\usepackage{booktabs,graphicx}
\usepackage{enumitem}
\usepackage{tikz}
\usetikzlibrary{arrows.meta,calc,positioning}

\newif\iftcoloravailable
\IfFileExists{tcolorbox.sty}{
  \tcoloravailabletrue
  \usepackage[most]{tcolorbox}
}{
  \tcoloravailablefalse
  \usepackage{framed}
  \newenvironment{tcolorbox}[1][]{\begin{framed}\small}{\end{framed}}
}
\IfFileExists{fvextra.sty}{
  \usepackage{fvextra}
}{
  \usepackage{listings}
  \lstnewenvironment{Verbatim}[1][]{
    \lstset{basicstyle=\ttfamily\footnotesize,breaklines=true,
      columns=fullflexible,keepspaces=true,showstringspaces=false}
  }{}
}

\iftcoloravailable
\tcbset{
  promptbox/.style={
    enhanced,
    width=\textwidth,
    colback=gray!3,
    colframe=black!40,
    colbacktitle=gray!55,
    coltitle=white,
    boxrule=0.4pt,
    arc=1mm,
    left=2mm,
    right=2mm,
    top=1.2mm,
    bottom=1.2mm,
    fonttitle=\bfseries\small
  }
}
\fi

\usepackage[T1]{fontenc}

\usepackage[utf8]{inputenc}

\usepackage{microtype}

\IfFileExists{inconsolata.sty}{\usepackage{inconsolata}}{\usepackage{courier}}

\title{LLMs Struggle to Measure What Distinguishes Students of Different
Proficiency Levels: A Study of Item Discrimination in Reading Comprehension Assessment}

\author{
  \textbf{Han Chen}\textsuperscript{1,*},
  \textbf{Ming Li}\textsuperscript{1,2,*},
  \textbf{Chenguang Wang}\textsuperscript{3},
  \textbf{Yijun Liang}\textsuperscript{2},
  \textbf{Dawei Zhou}\textsuperscript{3},
  \textbf{Hong Jiao}\textsuperscript{2}, 
  \textbf{Tianyi Zhou}\textsuperscript{1}\\ 
  \textsuperscript{1}MBZUAI ~~~~
  \textsuperscript{2}University of Maryland ~~~~
  \textsuperscript{3}Virginia Tech \\
  \texttt{\{minglii, hjiao\}@umd.edu} ~~~~
  \texttt{\{han.chen, tianyi.zhou\}@mbzuai.ac.ae} \\
  \textbf{Repo:} \url{https://github.com/MingLiiii/Item_Discrimination_Alignment}
}

\iftcoloravailable
\tcbset{
  finding style/.style={
    enhanced,
    colback=gray!10,           
    colframe=gray!60,         
    boxrule=0.8pt,             
    arc=2mm,                   
    left=1mm, right=1mm,       
    top=1mm, bottom=1mm,
    before skip=10pt, after skip=10pt,
    fontupper=\normalfont,     
    fonttitle=\normalfont\bfseries, 
  }
}

\newtcolorbox{findingbox}[1][]{finding style,#1}
\else

\fi

\begin{document}
\maketitle

\begingroup
\renewcommand{\thefootnote}{*}
\footnotetext{Equal contribution.}
\endgroup

\begin{abstract}

Existing work on LLM-based educational assessment has focused largely on item difficulty, but difficulty alone does not indicate whether an item meaningfully distinguishes higher- from lower-proficiency students. Item discrimination captures this complementary and fundamental psychometric property. We investigate whether LLMs can predict human item discrimination from assessment content. We evaluate 42 proprietary and open-weight LLMs using two complementary approaches. Direct discrimination prediction asks models to explicitly predict an item's discrimination value, while response-based proxy estimation treats LLM answers as synthetic responses and applies a Classical Test Theory (CTT)-inspired item-rest calculation. Direct predictions show weak alignment with human item discrimination. The response-based proxy provides a stronger but still limited ranking signal, reaching a CEFR-stratified rank correlation of 0.231. Further analysis shows that this correlation comes mainly from differences across models rather than proficiency prompts that reliably simulate students at different ability levels. Current LLMs therefore contain some discrimination-relevant information, but they do not yet reliably model the ability-conditioned human response behavior that gives item discrimination its psychometric meaning.

\end{abstract}

\section{Introduction}
\label{sec:introduction}

Item discrimination is a core property in educational measurement: it quantifies how well an assessment item distinguishes higher- from lower-proficiency examinees within a specified population. In classical test theory (CTT), it is commonly measured as the correlation between item correctness and total score~\citep{lord2008statistical,crocker1986introduction,moses2017review}. High values indicate that stronger examinees are more likely to answer correctly. By contrast, low or negative values indicate weak or reversed separation~\citep{ebel1972essentials,moses2017review,mccowan1999item}. Discrimination is therefore crucial for test construction because equally difficult items can differ substantially in diagnostic value~\citep{eignor2013standards,haladyna2013developing}.

Traditionally, item discrimination is estimated from human response data collected through large-scale pretesting, which makes calibration costly and time-consuming for newly developed items~\citep{crocker1986introduction,moses2017review,mccowan1999item}. The emergence of Large Language Models (LLMs) offers a potential alternative, as they can process assessment items and provide scalable judgments about item quality and difficulty~\citep{li2025item,veeramani2024large,li2025can}. However, prior work has focused largely on item difficulty, operationalized either through response proportions or model-based parameters such as Rasch difficulty~\citep{alkhuzaey2024text,yaneva2024findings,li2025can}. Whether LLMs can predict the distinct, ability-differentiating signal of discrimination remains less clear.

We study whether current LLMs can estimate human-calibrated item discrimination from reading-comprehension item content in a zero-shot setting. This question is difficult to study at scale because public resources rarely pair item content with discrimination values derived from real examinee responses. We therefore use the Cambridge Multiple-Choice Questions Reading Dataset~\citep{mullooly2023cambridge}, one of the few available datasets containing reading items and item-level psychometric statistics~\citep{liusie2023analysis}. It allows direct comparison between LLM estimates and human-calibrated discrimination.

We evaluate two complementary approaches: \textbf{Direct discrimination prediction} asks an LLM to estimate the discrimination value from the complete item. This approach tests whether the model can explicitly judge how well an item separates higher- and lower-proficiency examinees~\citep{crocker1986introduction,moses2017review}. \textbf{Response-based proxy estimation} instead asks LLMs to answer items without access to the answer key. We then compute an item-rest association from their correctness patterns, following the CTT principle that discrimination is expressed through covariance between item success and respondent capability~\citep{crocker1986introduction,moses2017review,sauberli2025llms,maeda2025field}. Because the release provides neither original test forms nor human response matrices, this is a proficiency-stratified synthetic proxy rather than a direct estimate of the original human statistic.

In this work, the zero-shot setting provides the primary test of direct discrimination prediction, while additional analyses clarify the resulting performance. Few-shot demonstrations test whether examples improve interpretation of the prediction scale, while task-grouped supervised text baselines ask whether item content supports discrimination prediction when labels are available. For response-based proxy estimation, we examine proficiency-prompt ordering, single-model versus heterogeneous response pools, score balancing, and ability-conditioned distractor behavior to identify which variation in LLM responses produces the predictive signal.

Our evaluation reflects two properties of discrimination. First, because the statistic is population-conditioned, overall performance can conflate average differences across proficiency groups with item ranking within a target group. We report both. Second, the response-based proxy can correlate with human discrimination through cross-model covariance without reflecting the human proficiency variation that defines the construct. We therefore separate useful ranking from evidence that a synthetic pool represents the intended examinee population. Generic low-, medium-, and high-proficiency prompts are treated as testable proficiency manipulations, not assumed human groups.

Our results support three claims. \textbf{First}, direct discrimination prediction does not reliably predict human item discrimination. \textbf{Second}, a CTT-style item-rest proxy computed from heterogeneous LLM responses reaches a within-CEFR rank correlation of 0.231, outperforming direct discrimination prediction but remaining too weak for item calibration. \textbf{Third}, the proxy's predictive signal comes mainly from differences across LLMs, rather than from proficiency prompts that reliably simulate ordered human ability levels. Direct predictions, response-based proxies, and pool ablations provide the primary evidence. The other analyses serve as diagnostics and robustness checks.

Our contributions are threefold. (1) We provide a large-scale evaluation of whether LLMs can predict human-calibrated item discrimination, comparing direct discrimination prediction and response-based proxy estimation across 42 models and 168 model-prompt configurations. (2) We extend LLM-based item analysis beyond difficulty and use a population-aware evaluation that separates differences between CEFR levels from item ranking within each level. (3) We identify the source of the item-rest proxy's correlation with human discrimination, showing that it comes mainly from cross-model variation rather than proficiency prompts that simulate ordered human ability levels, and clarify the implications for item-screening workflows.

\section{Related Work}
Most automated item-quality research predicts item difficulty from response or textual features, with recent work extending this direction to LLM-based signals~\citep{loukina2016textual,benedetto2023quantitative,li2025item, feng2025reasoning,peters2025text}. Difficulty and discrimination are distinct: a hard item need not distinguish examinees by proficiency, and LLM-based estimation of human item discrimination remains comparatively understudied~\citep{demars2010item,lord2012applications, han-etal-2025-leveraging-fine, wang2026cognitive}.

Synthetic LLM responses have also been explored for student simulation,
virtual pretesting, difficulty estimation, and psychometric item evaluation
\citep{markel2023gpteach,park2024large,liu2025leveraging},
although prior studies report mismatches with human error distributions
and ability-conditioned response patterns. We instead evaluate against released human item discrimination and distinguish differences between CEFR levels from item ranking within each level. A complete review is provided in Appendix~\ref{sec:related-work}.

\section{Data and Evaluation Protocol}

\subsection{Dataset and Task Description}

We study item-level discrimination prediction for multiple-choice reading comprehension assessment. Given an item, the goal is to predict how well an item distinguishes higher- and lower-proficiency test-takers. We use the Cambridge Multiple-Choice Questions Reading Dataset~\cite{mullooly2023cambridge} and the target label is the human-calibrated item discrimination value obtained from student pretesting. Because discrimination is population-dependent, we use each item's Common European Framework of Reference for Languages (CEFR) level (B1 to C2) as the finest observed target-population stratum available in the released data. CEFR does not fully identify an examinee population: unobserved cohort, background, and administration differences may remain. The distributed JSON file used in our experiments contains all 793 item records from 120 Cambridge reading-comprehension tasks. Each item includes the passage, question stem, four answer options, the correct answer, and psychometric statistics, including scaled Rasch difficulty and CTT-derived discrimination. Let
\[
\mathcal{D}=\{(x_i,a_i^*,y_i)\}_{i=1}^{N}
\]
denote the dataset, where $x_i$ is the full item context, including the passage, question, and answer options; $a_i^*$ is the ground-truth answer; and $y_i$ is the human-derived item discrimination label.

Under Classical Test Theory (CTT), discrimination is computed as the point-biserial correlation between item correctness and total test score:
\[
y_i = \mathrm{corr}(v_i,s),
\]
where $v_i\in\{0,1\}$ indicates whether an examinee answers item $i$ correctly, and $s$ denotes the examinee's total test score. A larger positive value indicates better discrimination, as higher-scoring examinees are more likely to answer the item correctly, whereas values near zero suggest weak separation between high- and low-scoring examinees. This released value is our \emph{human target statistic}. Its original administration-specific score, test form, and response matrix are not available. Response-based proxy estimation instead constructs a \emph{synthetic CEFR-wide capability score} from LLM answers and measures empirical alignment between the resulting proxy and the released target. The two quantities share an association-based form but need not have the same psychometric interpretation.
Table~\ref{tab:population} summarizes the dataset and psychometric targets.
\begin{table}[t]
\centering
\small
\resizebox{\linewidth}{!}{
\begin{tabular}{lrrrrr}
\toprule
CEFR & Items & Tasks & Scaled Rasch diff. & Disc. mean & Disc. SD \\
\midrule
B1 & 140 & 28 & 60.95 & 0.414 & 0.134 \\
B2 & 422 & 58 & 64.33 & 0.336 & 0.128 \\
C1 & 169 & 25 & 73.18 & 0.304 & 0.105 \\
C2 &  62 &  9 & 78.15 & 0.239 & 0.134 \\
\midrule
All & 793 & 120 & 66.70 & 0.335 & 0.133 \\
\bottomrule
\end{tabular}}
\caption{Population structure of the Cambridge JSON file used in our experiments. Difficulty is the released scaled Rasch value, not proportion correct. Human-calibrated discrimination differs substantially across CEFR strata, motivating both within-CEFR and overall evaluation.}
\label{tab:population}
\end{table}

We evaluate two LLM-based approaches. In \textbf{direct discrimination prediction}, the model receives the item context and ground-truth answer, then outputs a single discrimination score. We provide the answer key because discrimination concerns whether selecting the keyed option covaries with examinee proficiency, and item developers ordinarily know the intended answer when screening a new item. This also lets the model compare the keyed response with the distractors without conflating discrimination estimation with solving accuracy. In \textbf{response-based proxy estimation}, LLMs answer each item without access to the ground-truth answer, and each scored response vector serves as a synthetic respondent. Motivated by CTT item analysis, we correlate item correctness with a same-CEFR synthetic rest score. This proxy is related to, but does not reconstruct, the released human item-total statistic; Section~\ref{sec_ctt_calibration} defines it precisely.

\begin{table*}[ht]
\centering
\small
\resizebox{0.9\textwidth}{!}{
\begin{tabular}{lrrrrrrr}
\toprule
& & & \multicolumn{2}{c}{Overall} &
\multicolumn{3}{c}{Within CEFR} \\
\cmidrule(lr){4-5}\cmidrule(lr){6-8}
Family & Models & Configs. & Median $\rho$ & Best $\rho$ in family &
Median $\rho$ & Best $\rho$ in family & Positive 95\% CIs \\
\midrule
GPT & 8 & 32 & $-0.071$ & \textbf{0.152} & 0.026 & \textbf{0.172} & 12.5\% \\
Claude & 2 & 8 & $-0.124$ & 0.007 & 0.000 & 0.067 & 12.5\% \\
Gemini & 3 & 12 & $-0.034$ & 0.013 & 0.014 & 0.087 & 16.7\% \\
Llama & 6 & 24 & $-0.019$ & 0.064 & $-0.008$ & 0.057 & 0.0\% \\
Qwen & 16 & 64 & $-0.018$ & 0.112 & 0.025 & 0.102 & 9.4\% \\
Other open & 7 & 28 & $-0.019$ & 0.070 & $-0.013$ & 0.061 & 0.0\% \\
\midrule
All & 42 & 168 & $-0.035$ & \textbf{0.152} & 0.012 & \textbf{0.172} & 7.7\% \\
\bottomrule
\end{tabular}}
\caption{Distribution of zero-shot direct-prediction results over all 168 model-prompt configurations. Overall $\rho$ ranks all items together; within-CEFR $\rho$ uses only within-stratum ranks. ``Best'' is the post-hoc maximum within each family. The final column is the percentage of configurations whose 95\% task-clustered bootstrap interval excludes zero on the positive side.}
\label{tab:direct_families}
\end{table*}

\subsection{Prompted Respondent Profiles}
\label{sec_prompted_profiles}

To test whether generic proficiency prompts can elicit proficiency-dependent item behavior, we evaluate each model under four configurations:
\[
\mathcal{P} = \{p_0, p_{\text{low}}, p_{\text{mid}}, p_{\text{high}}\}.
\]
The default configuration $p_0$ uses no explicit student role. For $p_{\text{low}}$, $p_{\text{mid}}$, and $p_{\text{high}}$, the prompt asks the LLM to act as a lower-, medium-, or higher-proficiency Cambridge English test-taker, respectively. We deliberately use minimal proficiency descriptions rather than specifying error types or response strategies. Richer prompts could more directly induce lower accuracy. However, they would conflate the model's interpretation of proficiency with prompt-imposed error behavior. Our goal is to test whether LLMs translate coarse proficiency levels into ordered, human-aligned response patterns without detailed behavioral constraints.

For direct discrimination prediction, these prompted profiles are not intended as valid psychometric estimators. They are a prompt-sensitivity analysis, motivated by prior proficiency-prompting work, that tests whether eliciting different examinee perspectives changes item-level numerical judgments. The no-persona condition remains the primary setting. For response-based proxy estimation, they provide candidate proficiency-conditioned response vectors. We subsequently test whether the prompts induce ordered accuracy and item-level sensitivity aligned with human item discrimination. Detailed prompts are in Appendix~\ref{app:prompts}.

\subsection{Implementation Details}

\paragraph{Models and inference.}

We evaluate 42 proprietary and open-weight LLMs spanning the GPT, Claude, Gemini, Llama, Mistral, OLMo, Phi, and Qwen families and a broad capability range. Appendix~\ref{app:model_details} lists the evaluated models and summarizes the inference and output-parsing conventions used in our experiments.

\paragraph{Evaluation metrics.}

We report \textbf{Spearman correlation} for item ranking and \textbf{RMSE} for numerical calibration. The main analyses use overall correlation across all items and a CEFR-stratified Spearman correlation, which we call within-CEFR correlation. We rank predictions and labels separately within each CEFR stratum, standardize the two rank vectors within that stratum, and then pool the standardized ranks. The resulting statistic is the item-weighted mean of the four within-level Spearman correlations and contains no comparisons between items from different CEFR levels. Separate B1, B2, C1, and C2 correlations and their unweighted macro average are reported in Appendix Table~\ref{tab:within_cefr_detail}. All fitted supervised predictors use training folds only, and evaluation remains out-of-fold.

All main confidence intervals use 1,000 bootstrap samples with the 120 reading tasks as clusters. The zero-shot prompts were fixed before evaluation and were not selected using a development split or optimized against discrimination labels.

\section{Direct Discrimination Prediction}
\label{sec_direct_prediction}

\subsection{Direct Discrimination Prediction Fails}

\paragraph{Claim 1: Direct discrimination prediction does not reliably predict human item discrimination.} Across 168 configurations, direct discrimination prediction does not consistently predict human item discrimination. Table~\ref{tab:direct_families} summarizes the zero-shot results by model family. Complete model-level results are in Appendix~\ref{app:complete_results}. Most overall correlations are near zero or negative. However, when prediction and target ranks are compared only within CEFR levels, the negative correlations generally move toward zero. This shows that they mainly reflect differences across CEFR levels, rather than a consistent tendency to reverse the ranking of items within the same level.

Median within-CEFR $\rho$ is 0.012 (IQR $-0.015$ to 0.044), and only 13 configurations have task-clustered intervals above zero. The best configuration reaches within-CEFR $\rho=0.172$ (95\% CI [0.109, 0.242]). However, the correlation remains low, and the configuration was selected after comparing 168 alternatives. General language ability therefore does not reliably translate into estimates of human item discrimination.

\begin{figure}[t]
    \centering
    \includegraphics[width=0.92\linewidth]{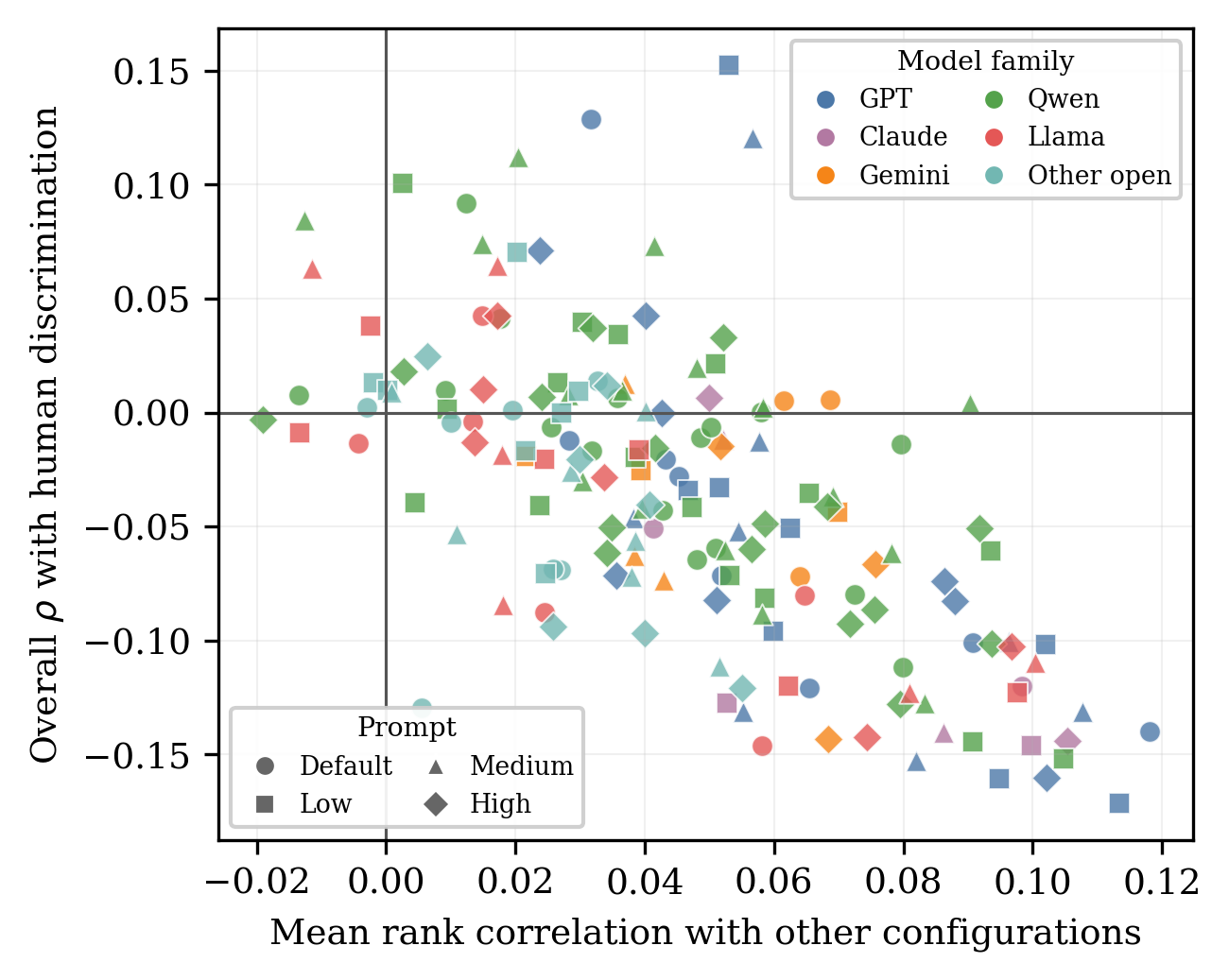}
    \caption{Direct predictions show only weak inter-model agreement. Each point is one model-prompt configuration; colors denote the same model family groups, and markers denote prompt conditions. The x-axis is the prediction vector's mean Spearman correlation with all other configurations, and the y-axis is its overall rank correlation with human discrimination. Even the most mutually aligned configurations have mean model-side agreement of only about 0.12.}
    \label{fig:consensus_alignment}
\end{figure}

Figure~\ref{fig:consensus_alignment} further shows that agreement among direct predictions is weak: no configuration has mean rank correlation above approximately 0.12 with the others. Configurations with relatively greater agreement tend to be more negatively aligned with human discrimination in the overall evaluation, but this association largely collapses under the within-CEFR evaluation (Appendix Figure~\ref{fig:consensus_population_adjustment}). The limited shared structure is consistent with common length- and difficulty-related heuristics, not reliable reverse ranking. Among the 19 models with negative overall direct correlations, predictions correlate on average with passage length at 0.340 and scaled human Rasch difficulty at 0.180. By contrast, for the two positively aligned models, the corresponding averages are 0.028 and 0.000. Since scaled Rasch difficulty and discrimination correlate at $-0.374$, such heuristics can produce weak model-side similarity without predicting within-population discrimination. Population-description ablations, prompted respondent profiles, and prompt-condition ensembling likewise provide no stable gains (Appendix~\ref{app:population_aware}).

\subsection{Robustness: Few-Shot Calibration}

One explanation for the weak zero-shot results is that LLMs may not know the numerical range of human discrimination values or how to map their judgments onto that scale. We therefore provide labeled examples to test whether few-shot demonstrations improve scale interpretation and item ranking. We use task-disjoint, CEFR-matched demonstrations for three models, plus an eight-shot follow-up for two. All comparisons share the same no-persona targets and metadata.

Table~\ref{tab:fewshot_main} reports the zero- and few-shot comparison.
\begin{table}[t]
\centering
\small
\setlength{\tabcolsep}{3.5pt}
\resizebox{\columnwidth}{!}{%
\begin{tabular}{llrrr}
\toprule
Model & Setting & Overall $\rho$ & Within CEFR $\rho$ & RMSE \\
\midrule
GPT-5.5 & Zero-shot & 0.129 & 0.103 & 0.154 \\
        & Four-shot mean & 0.263 & 0.038 & 0.141 \\
        & Eight-shot & 0.261 & 0.013 & 0.138 \\
\midrule
GPT-4.1 & Zero-shot & $-0.140$ & 0.053 & 0.193 \\
        & Four-shot mean & 0.176 & $-0.012$ & 0.144 \\
\midrule
Qwen3-32B & Zero-shot & $-0.043$ & 0.025 & 0.211 \\
           & Four-shot mean & 0.231 & 0.010 & 0.140 \\
           & Eight-shot & 0.296 & 0.057 & 0.142 \\
\bottomrule
\end{tabular}}
\caption{Few-shot direct discrimination prediction. Four-shot values average the evaluation metrics across three independently sampled, task-disjoint, CEFR-matched demonstration sets. Demonstrations improve scale calibration but not within-CEFR ranking.}
\label{tab:fewshot_main}
\end{table}

All nine four-shot runs improve overall correlation and reduce RMSE, but performance remains limited. No model improves its mean within-CEFR correlation across the three demonstration sets, and the four-shot mean within-CEFR correlations range only from $-0.012$ to 0.038. In addition, 98 to 100\% of predictions fall in $[0.2,0.6]$. The eight-shot follow-up also yields no reliable within-population gain. Under this protocol, demonstrations mainly teach the output range and CEFR-level scale rather than item ranking.

\section{Response-Based Proxy Estimation}
\label{sec:response_based}

\subsection{Synthetic Item-Rest Proxy Formulation}
\label{sec_ctt_calibration}
Section~\ref{sec_direct_prediction} asked LLMs to predict discrimination directly. We now derive a response-based proxy from their answer patterns. Each model-prompt configuration produces a response vector across items and serves as a synthetic respondent. Following the CTT item-rest procedure, we estimate each item's discrimination by correlating correctness on that item with synthetic respondents' scores on the remaining items. Because discrimination is population-conditioned, each rest score uses only items from the target item's CEFR level. CEFR is the finest available population stratum, but the resulting statistic remains a synthetic proxy rather than the original human estimate.

Let $\mathcal{M}$ denote the set of LLMs and let $\mathcal{P}$ denote the four proficiency configurations defined in Section~\ref{sec_prompted_profiles}. For each item $i$, model $m\in\mathcal{M}$ under configuration $p\in\mathcal{P}$ predicts an answer $\hat a_{i,m,p}$, which we convert into a binary correctness value:
\[
v_{i,m,p}=\mathbb{I}(\hat a_{i,m,p}=a_i^*).
\]
For computing the proxy, each model-configuration pair $(m,p)$ serves as one response vector in the synthetic pool. We construct one pool for each configuration, as well as a full heterogeneous pool:
\[
\begin{aligned}
\mathcal R_p
  &= \{(m,p): m\in\mathcal M\},
     && p\in\mathcal P,\\
\mathcal R_{\mathrm{all}}
  &= \mathcal M\times\mathcal P .
\end{aligned}
\]

Let $c_i$ denote item $i$'s CEFR level. For a given respondent pool $\mathcal{R}_s$, where $s\in\mathcal{P}\cup\{\mathrm{all}\}$, the CEFR-stratified total score is
\[
T^{(s,\ell)}_{m,p}=\sum_{j:c_j=\ell} v_{j,m,p}.
\]
Our primary score excludes the focal item,
\[
T^{(s,\ell)}_{m,p,-i}
  =\sum_{\substack{j:c_j=\ell\\j\neq i}}v_{j,m,p},
\]
and the CEFR-stratified synthetic item-rest estimate is
\[
\begin{aligned}
\tilde{y}^{(s)}_i
&= \operatorname{corr}\Bigl(
\{v_{i,m,p}\}_{(m,p)\in\mathcal{R}_s},\\
&\hspace{3.7em}
\{T^{(s,c_i)}_{m,p,-i}\}_{(m,p)\in\mathcal{R}_s}
\Bigr).
\end{aligned}
\]

If all response vectors in a pool have the same correctness value for an item, the correlation is undefined. We assign it zero because the item does not distinguish vectors in that pool. Excluding the focal item avoids part-whole inflation. We retain the uncorrected item-total definition only as a comparability analysis because the released Cambridge description does not provide enough information to determine its exact implementation (Appendix~\ref{app:population_aware}).

Crucially, the released items span 120 reading tasks and are not identified as a single test form. Thus $T_{m,p,-i}^{(s,c_i)}$ ranks synthetic respondents by CEFR-wide LLM capability. It is neither the unknown score used in human pretesting nor evidence that the LLM pool is a matched human population. We evaluate only its empirical alignment with the released human target.

For the null comparison, we preserve coarse item facility by permuting human discrimination only among items in the same CEFR level and synthetic-accuracy decile. Appendix Table~\ref{tab:ctt_null} reports this accuracy-conditioned comparison.

\subsection{The Response-Based Proxy Provides Limited Ranking Signal}

\paragraph{Claim 2: The response-based item-rest proxy improves within-CEFR ranking over direct discrimination prediction but remains weak.} Our primary empirical result appears in Table~\ref{tab:table2}.
Every respondent pool used for response-based proxy estimation produces positive overall and within-CEFR correlations. The primary corrected full heterogeneous pool reaches a within-CEFR correlation of $\rho=0.231$, improving over direct discrimination prediction but remaining too weak to replace human pretesting. Pool ablations in Section~\ref{sec:diagnosis} show that this correlation comes primarily from variation across models rather than ordered variation across proficiency prompts. An accuracy-conditioned null shuffles human discrimination only among same-CEFR items in the same synthetic-accuracy decile; none of 1,000 replicates reaches the observed correlation ($p=0.001$). Thus, the ranking is not explained by coarse synthetic item accuracy alone, although the null does not establish population-matched measurement. Complete results are reported in Table~\ref{tab:table2} and Appendix Table~\ref{tab:ctt_null}.

\begin{table}[t]
\centering
\resizebox{\linewidth}{!}{
\begin{tabular}{lccc}
\toprule
Pool or reference & RMSE $\downarrow$ & Overall $\rho$ & Within-CEFR $\rho$ \\
\midrule
\multicolumn{4}{l}{\emph{Descriptive references}} \\
Global mean & 0.133 & \textemdash & \textemdash \\
Within-CEFR means & 0.125 & \textemdash & \textemdash \\
\midrule
\multicolumn{4}{l}{\emph{Corrected item-rest}} \\
No-persona & 0.284 & 0.216 & 0.210 \\
Low prompt & 0.281 & 0.242 & 0.218 \\
Medium prompt & 0.286 & 0.211 & 0.187 \\
High prompt & 0.280 & 0.243 & 0.214 \\
\textbf{Full pool} & \textbf{0.270} & \textbf{0.255} & \textbf{0.231} \\
\midrule
\multicolumn{4}{l}{\emph{Uncorrected item-total comparator}} \\
Full pool & 0.275 & 0.249 & 0.230 \\
\bottomrule
\end{tabular}}
\caption{Results for response-based proxy estimation using CEFR-stratified synthetic scores. Corrected item-rest is the primary estimator, and uncorrected item-total is shown for comparison. The full pool combines all 42 models and four prompt conditions. Within-CEFR $\rho$ ranks and standardizes predictions and labels separately within each CEFR stratum before pooling. Additional protocol comparisons are in Appendix~\ref{app:population_aware}.}
\label{tab:table2}
\end{table}

\paragraph{Calibration limits and robustness.} Numerical calibration remains poor: the corrected item-rest proxy has RMSE 0.270, compared with 0.133 for the global-mean descriptive reference and 0.125 for the within-CEFR-mean descriptive reference. The two mean references are computed from the gold discrimination labels, so their low RMSE is descriptive and cannot be achieved when predicting new items without labels. The response-based proxy has substantially larger numerical error, even though its positive within-CEFR correlation indicates some ability to rank items. It therefore provides limited ranking information but does not predict discrimination values accurately.

The uncorrected CEFR-wide item-total estimator yields overall $\rho=0.249$ and within-CEFR $\rho=0.230$. The corrected estimator slightly improves overall alignment and RMSE without changing the substantive conclusion. The historical global-score protocol yields overall $\rho=0.243$ on the audited response matrix. The full protocol comparison and parser conventions appear in Appendix~\ref{app:population_aware}. CEFR stratification and focal-item correction therefore improve the formulation without creating the positive response-based result.

The result is also stable when the synthetic capability score is built from random, test-length-like subsets rather than every same-CEFR item. Across 200 draws, mean within-CEFR $\rho$ is 0.227, 0.229, and 0.230 for 20-, 40-, and 80-item rest pools, respectively (Appendix Table~\ref{tab:ctt_pool_sensitivity}). This sensitivity analysis addresses score length, while the original human test forms remain unavailable.

\subsection{Screening Diagnostic}

We also evaluate whether predicted scores retrieve the top or bottom $k\%$ of human-calibrated items using Overlap@$k$ ($k=10,20$). The CEFR-stratified item-rest proxy is most useful for low-discrimination screening: the corrected full-pool proxy retrieves 36.7\% and 37.7\% of bottom items at @10 and @20, respectively, whereas direct predictions remain close to random. Because even the best setting retrieves fewer than half of the target items, this signal supports triage for human review rather than full calibration. At a 20\% review budget, the corrected proxy retrieves 40.8\% of items whose human discrimination is at most 0.2, with 26.4\% precision and $2.03\times$ enrichment over their dataset prevalence. Task-clustered uncertainty, metric definitions, and complete results appear in Appendix~\ref{app:complete_results}.

\begin{table*}[t]
\centering
\small
\setlength{\tabcolsep}{4.5pt}
\begin{tabular}{lccrrr}
\toprule
Estimator & \shortstack{Prediction-time\\input} &
\shortstack{Disc.\ labels\\in training?} & Overall $\rho$ &
Within-CEFR $\rho$ & \shortstack{Within-CEFR\\95\% CI} \\
\midrule
\multicolumn{6}{l}{\emph{Label-trained predictors without human response statistics}} \\
CEFR-mean prior & CEFR level & Yes & 0.266 & \textemdash & \textemdash \\
TF-IDF ridge & Item + key & Yes & 0.215 & $-0.039$ & [$-0.111$, 0.042] \\
Frozen-embedding ridge & Item + key & Yes & 0.158 & $-0.076$ & [$-0.149$, $-0.003$] \\
Fine-tuned ModernBERT & Item + key & Yes & 0.215 & 0.014 & [$-0.067$, 0.091] \\
\midrule
\multicolumn{6}{l}{\emph{No discrimination-label training}} \\
Post-hoc best direct configuration & Item + key + CEFR & No & 0.152 & 0.172 & [0.109, 0.242] \\
CEFR item-rest proxy & LLM answers + key & No & 0.255 & 0.231 & [0.168, 0.300] \\
\midrule
\multicolumn{6}{l}{\emph{Human-pretest benchmark}} \\
Rasch difficulty $\rightarrow$ discrimination & Human Rasch diff. & Yes & 0.370 & 0.246 & [0.173, 0.314] \\
\bottomrule
\end{tabular}
\caption{Inputs and training signals for discrimination prediction. The CEFR-mean prior has no item-specific ranking within a CEFR level, so its within-CEFR entries are marked with dashes. The Rasch-difficulty row learns its mapping on outer training folds and applies it to held-out items. Because its input is estimated from human pretesting, it is a benchmark rather than a deployable alternative. The direct row is GPT-5.5 Low, selected descriptively from 168 configurations; CIs are task-clustered.}
\label{tab:supervised_main}
\vspace{-4mm}
\end{table*}

\section{Sources of Predictive Signal}
\label{sec:diagnosis}

Sections~\ref{sec_direct_prediction} and~\ref{sec:response_based} show that direct discrimination prediction provides little reliable within-population ranking, whereas the response-based item-rest proxy reaches a within-CEFR correlation of 0.231. We now compare the information available to different estimators and identify which variation in the synthetic response pool produces this correlation.

\subsection{Discrimination-Label Training and Information Controls}

Table~\ref{tab:supervised_main} separates the information available at prediction time from the labels used for training. The first block asks whether human discrimination labels can supervise prediction from CEFR or item content. We fit a CEFR-mean prior, TF-IDF ridge, frozen-embedding ridge, and ModernBERT using five-fold task-grouped, CEFR-stratified cross-validation. The text models receive the passage, question, options, and correct answer. CEFR is used for splitting and evaluation but is not an input feature.

The second block contains the two methods that use no discrimination labels: GPT-5.5 uses direct discrimination prediction, whereas the response-based item-rest proxy is computed from LLM answer correctness. The final row is deliberately different. It fits a one-feature ridge mapping from human scaled Rasch difficulty to human item discrimination on the training folds, then applies that mapping to held-out items. We include it as a human-pretest benchmark because the Rasch difficulty value already contains information from human responses.

The tested label-trained item-content models do not achieve the response-based proxy's within-CEFR correlation. The CEFR-mean prior reaches overall $\rho=0.266$ by capturing differences between CEFR levels, but it provides no item-specific ranking within a level; we therefore do not report a within-CEFR correlation for this prior. Supervised text models reach overall correlations of 0.158 to 0.215 while remaining near zero within CEFR. Under grouped out-of-fold evaluation, they mainly predict between-group differences rather than within-group item ranking. This diagnostic does not establish that item text contains no discrimination signal. GPT-5.5 Low is the best direct-prediction configuration among those evaluated, but it was selected post hoc and is not model-selection equivalent to the supervised rows. Per-CEFR results appear in Appendix Table~\ref{tab:within_cefr_detail}.

The Rasch-difficulty benchmark reaches overall $\rho=0.370$ and within-CEFR $\rho=0.246$. Its high value is expected because scaled Rasch difficulty and discrimination are derived from human pretesting and correlate at $-0.374$ overall. The training folds identify the inverse direction. This row shows how much ranking signal is available from one human response statistic, but it is not a deployable predictor for items without pretest data. The CEFR-stratified item-rest proxy reaches a comparable within-CEFR correlation of 0.231 without using human-calibrated discrimination labels or other human response statistics, although it requires the answer key. Heterogeneous response patterns therefore enable discrimination prediction that the tested label-trained item-content models do not achieve reliably.

\subsection{Attributing the Response-Based Proxy's Predictive Signal}

\paragraph{Claim 3: The proxy's predictive signal comes mainly from differences across models rather than from the proficiency prompts.} Prompt-ordering controls and pool ablations identify the respondent variation underlying the response-based item-rest proxy.

\paragraph{Respondent-pool ablations (primary evidence).} Under the corrected item-rest protocol, the cross-model no-persona pool reaches within-CEFR $\rho=0.210$, whereas the four prompt conditions of one model yield mean $\rho=-0.006$. Randomly retaining one prompt condition per model gives mean $\rho=0.211$, compared with 0.231 for the full pool. Equalizing the six model-family groups reduces the mean to 0.189, while leave-one-family-out estimates range from 0.169 to 0.253. Thus, cross-model variation dominates the signal, but its magnitude depends moderately on the composition of the evaluated model pool; complete ablations appear in Appendix~\ref{app:persona_validity}.

\paragraph{Prompt-induced ordering (supporting control).} Consistent with the pool ablation, deterministic proficiency prompts do not produce a reliable within-model ability ordering. Only 8 of 42 models satisfy $\mathrm{Acc}_{\mathrm{low}}<\mathrm{Acc}_{\mathrm{mid}}< \mathrm{Acc}_{\mathrm{high}}$, and the mean high versus low accuracy gap is only 0.56 percentage points. Item-level prompt-profile sensitivity is unrelated to human discrimination ($\rho=-0.039$, 95\% CI [$-0.113$, 0.039]). Per-model accuracies and error-overlap tests appear in Appendix Table~\ref{tab:answer_accuracy} and Appendix~\ref{app:persona_validity}.

\paragraph{Cross-model variation as the primary explanation.} Cross-model differences create positive item-rest covariance when stronger models answer both a focal item and the remaining items correctly more often. Partial overlap with human ability-based discrimination can then improve ranking without matching human proficiency groups or error processes. The ablations do not isolate capability from correlated cross-model differences such as architecture or training data.

\paragraph{Distractor behavior (supporting diagnostic).} Across 502 items, synthetic distractor endorsement moderately aligns with human endorsement ($\rho=0.315$, task-clustered 95\% CI [0.252, 0.369]), but synthetic and human distractor discrimination correlate at only 0.086 (95\% CI [0.032, 0.136]). Thus LLMs predict aggregate plausibility better than ability-conditioned attraction. Additional diagnostics appear in Appendix~\ref{app:error_distractor}.

\section{Conclusion}

We evaluated whether LLMs can predict item discrimination. Across 42 models, direct predictions align weakly with human labels, whereas response-based proxy estimates provide stronger but still limited ranking signal. That signal arises mainly from cross-model differences, not proficiency prompts that yield ordered, human-aligned behavior; it therefore does not validate the synthetic respondents as the target population. Item discrimination remains an open challenge for LLM-based assessment.

\clearpage
\section*{Limitations}

Our prompted-profile analysis uses generic deterministic instructions, so it does not rule out more structured student-error simulation. Future work could condition synthetic respondents on explicit epistemic states or reading-specific error mechanisms, including lexical matching, failure to locate evidence, incorrect inference integration, reference confusion, and attraction to plausible distractors~\citep{yuan2026towards,sonkar2024llm,miroyan2025parastudent,ross2025learning}. The key test would be whether these constraints produce ordered accuracy and ability-conditioned errors that align with human responses and thereby improve response-based proxy estimation. Even then, such simulations would supplement rather than replace human pretesting.

Moreover, our experiments are limited to reading-comprehension assessment. We adopt this scope because public datasets that pair item content with human-calibrated discrimination statistics are scarce. Thus, our results should be viewed as evidence from a rare available benchmark rather than a complete characterization of LLM-based discrimination estimation across assessment settings.

The released data do not identify the original test forms or provide human response matrices, so our CEFR-wide synthetic rest scores cannot reproduce the score or examinee population used for the released item-total statistic. The Cambridge items were also publicly available before many evaluated models were trained; because hosted-model training data are not auditable, we cannot rule out item or answer exposure. Finally, the family-balanced ablation shows that the response-based correlation depends partly on the composition of the model pool. These factors limit generalization to unseen items and new model collections.

\bibliography{custom}

@article{li2025can,
  title={Can LLMs Estimate Student Struggles? Human-AI Difficulty Alignment with Proficiency Simulation for Item Difficulty Prediction},
  author={Li, Ming and Chen, Han and Xiao, Yunze and Chen, Jian and Jiao, Hong and Zhou, Tianyi},
  journal={arXiv preprint arXiv:2512.18880},
  year={2025}
}

@article{mullooly2023cambridge,
  title={The cambridge multiple-choice questions reading dataset},
  author={Mullooly, Andrew and Andersen, {\O}istein and Benedetto, Luca and Buttery, Paula and Caines, Andrew and Gales, Mark JF and Karatay, Yasin and Knill, Kate and Liusie, Adian and Raina, Vatsal and others},
  year={2023},
  publisher={Cambridge University Press and Assessment}
}

@inproceedings{yaneva2024findings,
  title={Findings from the first shared task on automated prediction of difficulty and response time for multiple-choice questions},
  author={Yaneva, Victoria and North, Kai and Baldwin, Peter and Ha, Le An and Rezayi, Saed and Zhou, Yiyun and Choudhury, Sagnik Ray and Harik, Polina and Clauser, Brian},
  booktitle={Proceedings of the 19th Workshop on Innovative Use of NLP for Building Educational Applications (BEA 2024)},
  pages={470--482},
  year={2024}
}

@article{hurst2024gpt,
  title={Gpt-4o system card},
  author={Hurst, Aaron and Lerer, Adam and Goucher, Adam P and Perelman, Adam and Ramesh, Aditya and Clark, Aidan and Ostrow, AJ and Welihinda, Akila and Hayes, Alan and Radford, Alec and others},
  journal={arXiv preprint arXiv:2410.21276},
  year={2024}
}

@misc{openai_chatgpt_whisper_api_2023,
  author       = {{OpenAI}},
  title        = {Introducing APIs for {GPT-3.5 Turbo} and {Whisper}},
  year         = {2023},
  howpublished = {\url{https://openai.com/index/introducing-chatgpt-and-whisper-apis/}},
  note         = {Updated April 24, 2024; accessed August 1, 2026}
}

@misc{openai_gpt4o_mini_2024,
  author       = {{OpenAI}},
  title        = {{GPT-4o mini}: Advancing Cost-Efficient Intelligence},
  year         = {2024},
  month        = jul,
  howpublished = {\url{https://openai.com/index/gpt-4o-mini-advancing-cost-efficient-intelligence/}},
  note         = {Published July 18, 2024; accessed August 1, 2026}
}

@misc{openai_gpt4_1_2025,
  title        = {Introducing GPT-4.1 in the API},
  author       = {{OpenAI}},
  howpublished = {\url{https://openai.com/index/gpt-4-1/}},
  year         = {2025}
}

@misc{openai2025o3o4mini,
  author = {{OpenAI}},
  title = {Introducing OpenAI o3 and o4-mini},
  year = {2025},
  howpublished = {\url{https://openai.com/index/introducing-o3-and-o4-mini/}}
}

@misc{openai_gpt5_system_card_2025,
  title        = {GPT-5 System Card},
  author       = {{OpenAI}},
  howpublished = {\url{https://openai.com/index/gpt-5-system-card/}},
  year         = {2025}
}

@misc{anthropic2025claude37sonnet,
  title        = {Claude 3.7 Sonnet System Card},
  author       = {{Anthropic}},
  year         = {2025},
  howpublished = {\url{https://www.anthropic.com/claude-3-7-sonnet-system-card}}
}

@misc{google2024gemini20,
  title        = {Introducing Gemini 2.0: Our New AI Model for the Agentic Era},
  author       = {{Google DeepMind}},
  year         = {2024},
  howpublished = {\url{https://blog.google/innovation-and-ai/models-and-research/google-deepmind/google-gemini-ai-update-december-2024/}}
}

@article{comanici2025gemini,
  title={Gemini 2.5: Pushing the frontier with advanced reasoning, multimodality, long context, and next generation agentic capabilities},
  author={Comanici, Gheorghe and Bieber, Eric and Schaekermann, Mike and Pasupat, Ice and Sachdeva, Noveen and Dhillon, Inderjit and Blistein, Marcel and Ram, Ori and Zhang, Dan and Rosen, Evan and others},
  journal={arXiv preprint arXiv:2507.06261},
  year={2025}
}

@article{touvron2023llama,
  title={Llama 2: Open foundation and fine-tuned chat models},
  author={Touvron, Hugo and Martin, Louis and Stone, Kevin and Albert, Peter and Almahairi, Amjad and Babaei, Yasmine and Bashlykov, Nikolay and Batra, Soumya and Bhargava, Prajjwal and Bhosale, Shruti and others},
  journal={arXiv preprint arXiv:2307.09288},
  year={2023}
}

@article{grattafiori2024llama,
  title={The llama 3 herd of models},
  author={Grattafiori, Aaron and Dubey, Abhimanyu and Jauhri, Abhinav and Pandey, Abhinav and Kadian, Abhishek and Al-Dahle, Ahmad and Letman, Aiesha and Mathur, Akhil and Schelten, Alan and Vaughan, Alex and others},
  journal={arXiv preprint arXiv:2407.21783},
  year={2024}
}

@article{qwen2024qwen25,
  title   = {{Qwen2.5} Technical Report},
  author  = {{Qwen Team}},
  journal = {arXiv preprint arXiv:2412.15115},
  year    = {2024},
  doi     = {10.48550/arXiv.2412.15115}
}

@article{yang2025qwen3,
  title={Qwen3 technical report},
  author={Yang, An and Li, Anfeng and Yang, Baosong and Zhang, Beichen and Hui, Binyuan and Zheng, Bo and Yu, Bowen and Gao, Chang and Huang, Chengen and Lv, Chenxu and others},
  journal={arXiv preprint arXiv:2505.09388},
  year={2025}
}

@misc{qwen2026qwen35,
  author       = {{Qwen Team}},
  title        = {{Qwen3.5}: Towards Native Multimodal Agents},
  year         = {2026},
  month        = feb,
  howpublished = {\url{https://qwen.ai/blog?id=qwen3.5}},
  note         = {Published February 15, 2026; accessed August 1, 2026}
}

@article{abdin2024phi,
  title={Phi-4 technical report},
  author={Abdin, Marah and Aneja, Jyoti and Behl, Harkirat and Bubeck, S{\'e}bastien and Eldan, Ronen and Gunasekar, Suriya and Harrison, Michael and Hewett, Russell J and Javaheripi, Mojan and Kauffmann, Piero and others},
  journal={arXiv preprint arXiv:2412.08905},
  year={2024}
}

@misc{abdin2024phi3technicalreporthighly,
      title={Phi-3 Technical Report: A Highly Capable Language Model Locally on Your Phone}, 
      author={Marah Abdin and Jyoti Aneja and Hany Awadalla and Ahmed Awadallah and Ammar Ahmad Awan and Nguyen Bach and Amit Bahree and Arash Bakhtiari and Jianmin Bao and Harkirat Behl and Alon Benhaim and Misha Bilenko and Johan Bjorck and Sébastien Bubeck and Martin Cai and Qin Cai and Vishrav Chaudhary and Dong Chen and Dongdong Chen and Weizhu Chen and Yen-Chun Chen and Yi-Ling Chen and Hao Cheng and Parul Chopra and Xiyang Dai and Matthew Dixon and Ronen Eldan and Victor Fragoso and Jianfeng Gao and Mei Gao and Min Gao and Amit Garg and Allie Del Giorno and Abhishek Goswami and Suriya Gunasekar and Emman Haider and Junheng Hao and Russell J. Hewett and Wenxiang Hu and Jamie Huynh and Dan Iter and Sam Ade Jacobs and Mojan Javaheripi and Xin Jin and Nikos Karampatziakis and Piero Kauffmann and Mahoud Khademi and Dongwoo Kim and Young Jin Kim and Lev Kurilenko and James R. Lee and Yin Tat Lee and Yuanzhi Li and Yunsheng Li and Chen Liang and Lars Liden and Xihui Lin and Zeqi Lin and Ce Liu and Liyuan Liu and Mengchen Liu and Weishung Liu and Xiaodong Liu and Chong Luo and Piyush Madan and Ali Mahmoudzadeh and David Majercak and Matt Mazzola and Caio César Teodoro Mendes and Arindam Mitra and Hardik Modi and Anh Nguyen and Brandon Norick and Barun Patra and Daniel Perez-Becker and Thomas Portet and Reid Pryzant and Heyang Qin and Marko Radmilac and Liliang Ren and Gustavo de Rosa and Corby Rosset and Sambudha Roy and Olatunji Ruwase and Olli Saarikivi and Amin Saied and Adil Salim and Michael Santacroce and Shital Shah and Ning Shang and Hiteshi Sharma and Yelong Shen and Swadheen Shukla and Xia Song and Masahiro Tanaka and Andrea Tupini and Praneetha Vaddamanu and Chunyu Wang and Guanhua Wang and Lijuan Wang and Shuohang Wang and Xin Wang and Yu Wang and Rachel Ward and Wen Wen and Philipp Witte and Haiping Wu and Xiaoxia Wu and Michael Wyatt and Bin Xiao and Can Xu and Jiahang Xu and Weijian Xu and Jilong Xue and Sonali Yadav and Fan Yang and Jianwei Yang and Yifan Yang and Ziyi Yang and Donghan Yu and Lu Yuan and Chenruidong Zhang and Cyril Zhang and Jianwen Zhang and Li Lyna Zhang and Yi Zhang and Yue Zhang and Yunan Zhang and Xiren Zhou},
      year={2024},
      eprint={2404.14219},
      archivePrefix={arXiv},
      primaryClass={cs.CL},
      url={https://arxiv.org/abs/2404.14219}, 
}

@inproceedings{benedetto2023quantitative,
  title={A quantitative study of NLP approaches to question difficulty estimation},
  author={Benedetto, Luca},
  booktitle={International conference on artificial intelligence in education},
  pages={428--434},
  year={2023},
  organization={Springer}
}

@article{hsu2018automated,
  title={Automated estimation of item difficulty for multiple-choice tests: An application of word embedding techniques},
  author={Hsu, Fu-Yuan and Lee, Hahn-Ming and Chang, Tao-Hsing and Sung, Yao-Ting},
  journal={Information Processing \& Management},
  volume={54},
  number={6},
  pages={969--984},
  year={2018},
  publisher={Elsevier}
}

@book{demars2010item,
  title={Item response theory},
  author={DeMars, Christine},
  year={2010},
  publisher={Oxford University Press}
}

@book{hambleton1991fundamentals,
  title={Fundamentals of item response theory},
  author={Hambleton, Ronald K and Swaminathan, Hariharan and Rogers, H Jane},
  volume={2},
  year={1991},
  publisher={Sage}
}

@inproceedings{loukina2016textual,
  title={Textual complexity as a predictor of difficulty of listening items in language proficiency tests},
  author={Loukina, Anastassia and Yoon, Su-Youn and Sakano, Jennifer and Wei, Youhua and Sheehan, Kathy},
  booktitle={Proceedings of COLING 2016, the 26th International Conference on Computational Linguistics: Technical Papers},
  pages={3245--3253},
  year={2016}
}

@inproceedings{sano2015automated,
  title={Automated capturing of psycho-linguistic features in reading assessment text},
  author={Sano, Makoto},
  booktitle={annual meeting of the National Council on Measurement in Education, Chicago, IL},
  year={2015}
}

@inproceedings{huang2017question,
  title={Question Difficulty Prediction for READING Problems in Standard Tests},
  author={Huang, Zhenya and Liu, Qi and Chen, Enhong and Zhao, Hongke and Gao, Mingyong and Wei, Si and Su, Yu and Hu, Guoping},
  booktitle={Proceedings of the AAAI conference on artificial intelligence},
  volume={31},
  number={1},
  year={2017}
}

@inproceedings{devlin2019bert,
  title={Bert: Pre-training of deep bidirectional transformers for language understanding},
  author={Devlin, Jacob and Chang, Ming-Wei and Lee, Kenton and Toutanova, Kristina},
  booktitle={Proceedings of the 2019 conference of the North American chapter of the association for computational linguistics: human language technologies, volume 1 (long and short papers)},
  pages={4171--4186},
  year={2019}
}

@inproceedings{he2021automatically,
  title={Automatically predict question difficulty for reading comprehension exercises},
  author={He, Jun and Peng, Li and Sun, Bo and Yu, Lejun and Zhang, Yinghui},
  booktitle={2021 ieee 33rd international conference on tools with artificial intelligence (ictai)},
  pages={1398--1402},
  year={2021},
  organization={IEEE}
}

@article{li2025item,
  title={Item difficulty modeling using fine-tuned small and large language models},
  author={Li, Ming and Jiao, Hong and Zhou, Tianyi and Zhang, Nan and Peters, Sydney and Lissitz, Robert W},
  journal={Educational and Psychological Measurement},
  volume={85},
  number={6},
  pages={1065--1090},
  year={2025},
  publisher={SAGE Publications Sage CA: Los Angeles, CA}
}

@inproceedings{rogoz2024unibucllm,
  title={Unibucllm: Harnessing LLMs for automated prediction of item difficulty and response time for multiple-choice questions},
  author={Rogoz, Ana-Cristina and Ionescu, Radu Tudor},
  booktitle={Proceedings of the 19th Workshop on Innovative Use of NLP for Building Educational Applications (BEA 2024)},
  pages={493--502},
  year={2024}
}

@inproceedings{feng2025reasoning,
  title={Reasoning and sampling-augmented mcq difficulty prediction via llms},
  author={Feng, Wanyong and Tran, Peter and Sireci, Stephen and Lan, Andrew S},
  booktitle={International Conference on Artificial Intelligence in Education},
  pages={31--45},
  year={2025},
  organization={Springer}
}

@article{zotos2024you,
  title={Are you doubtful? Oh, it might be difficult then! Exploring the use of model uncertainty for question difficulty estimation},
  author={Zotos, Leonidas and van Rijn, Hedderik and Nissim, Malvina},
  journal={arXiv preprint arXiv:2412.11831},
  year={2024}
}

@article{chrysafiadi2013student,
  title={Student modeling approaches: A literature review for the last decade},
  author={Chrysafiadi, Konstantina and Virvou, Maria},
  journal={Expert Systems with Applications},
  volume={40},
  number={11},
  pages={4715--4729},
  year={2013},
  publisher={Elsevier}
}

@article{vanlehn1994applications,
  title={Applications of simulated students: An exploration},
  author={VanLehn, Kurt and Ohlsson, Stellan and Nason, Rod},
  journal={Journal of artificial intelligence in education},
  volume={5},
  pages={135--135},
  year={1994},
  publisher={AACE ASSOCIATION FOR THE ADVANCEMENT OF}
}

@article{graesser2005autotutor,
  title={AutoTutor: An intelligent tutoring system with mixed-initiative dialogue},
  author={Graesser, Arthur C and Chipman, Patrick and Haynes, Brian C and Olney, Andrew},
  journal={IEEE Transactions on Education},
  volume={48},
  number={4},
  pages={612--618},
  year={2005},
  publisher={IEEE}
}

@inproceedings{stasaski2020more,
  title={More diverse dialogue datasets via diversity-informed data collection},
  author={Stasaski, Katherine and Yang, Grace Hui and Hearst, Marti A},
  booktitle={Proceedings of the 58th annual meeting of the association for computational linguistics},
  pages={4958--4968},
  year={2020}
}

@inproceedings{park2023generative,
  title={Generative agents: Interactive simulacra of human behavior},
  author={Park, Joon Sung and O'Brien, Joseph and Cai, Carrie Jun and Morris, Meredith Ringel and Liang, Percy and Bernstein, Michael S},
  booktitle={Proceedings of the 36th annual acm symposium on user interface software and technology},
  pages={1--22},
  year={2023}
}

@article{kaser2024simulated,
  title={Simulated learners in educational technology: A systematic literature review and a turing-like test},
  author={K{\"a}ser, Tanja and Alexandron, Giora},
  journal={International Journal of Artificial Intelligence in Education},
  volume={34},
  number={2},
  pages={545--585},
  year={2024},
  publisher={Springer}
}

@article{corbett1994knowledge,
  title={Knowledge tracing: Modeling the acquisition of procedural knowledge},
  author={Corbett, Albert T and Anderson, John R},
  journal={User modeling and user-adapted interaction},
  volume={4},
  number={4},
  pages={253--278},
  year={1994},
  publisher={Springer}
}

@book{rasch1993probabilistic,
  title={Probabilistic models for some intelligence and attainment tests.},
  author={Rasch, Georg},
  year={1993},
  publisher={ERIC}
}

@book{lord2012applications,
  title={Applications of item response theory to practical testing problems},
  author={Lord, Frederic M},
  year={2012},
  publisher={Routledge}
}

@book{embretson2025item,
  title={Item response theory: Foundations for psychologists and social scientists},
  author={Embretson, Susan E and Reise, Steven P},
  year={2025},
  publisher={Routledge}
}

@inproceedings{markel2023gpteach,
  title={Gpteach: Interactive ta training with gpt-based students},
  author={Markel, Julia M and Opferman, Steven G and Landay, James A and Piech, Chris},
  booktitle={Proceedings of the tenth acm conference on learning@ scale},
  pages={226--236},
  year={2023}
}

@inproceedings{lee2023generative,
  title={Generative agent for teacher training: Designing educational problem-solving simulations with large language model-based agents for pre-service teachers},
  author={Lee, Unggi and Lee, Sanghyeok and Koh, Junbo and Jeong, Yeil and Jung, Haewon and Byun, Gyuri and Lee, Yunseo and Moon, Jewoong and Lim, Jieun and Kim, Hyeoncheol},
  booktitle={NeurIPS’23 Workshop on Generative AI for Education (GAIED)},
  year={2023}
}

@article{miroyan2025parastudent,
  title={ParaStudent: Generating and Evaluating Realistic Student Code by Teaching LLMs to Struggle},
  author={Miroyan, Mihran and Niousha, Rose and Gonzalez, Joseph E and Ranade, Gireeja and Norouzi, Narges},
  journal={arXiv preprint arXiv:2507.12674},
  year={2025}
}

@article{ross2025modeling,
  title={Modeling student learning with 3.8 million program traces},
  author={Ross, Alexis and Srivastava, Megha and Blanchard, Jeremiah and Andreas, Jacob},
  journal={arXiv preprint arXiv:2510.05056},
  year={2025}
}

@article{liu2025leveraging,
  title={Leveraging LLM respondents for item evaluation: A psychometric analysis},
  author={Liu, Yunting and Bhandari, Shreya and Pardos, Zachary A},
  journal={British Journal of Educational Technology},
  volume={56},
  number={3},
  pages={1028--1052},
  year={2025},
  publisher={Wiley Online Library}
}

@inproceedings{hayakawa2024can,
  title={Can LLMs solve reading comprehension tests as second language learners?},
  author={Hayakawa, Akio and Saggion, Horacio},
  booktitle={Fourth Workshop on Knowledge-infused Learning},
  year={2024}
}

@inproceedings{park2024large,
  title={Large language models are students at various levels: Zero-shot question difficulty estimation},
  author={Park, Jae-Woo and Park, Seong-Jin and Won, Hyun-Sik and Kim, Kang-Min},
  booktitle={Findings of the Association for Computational Linguistics: EMNLP 2024},
  pages={8157--8177},
  year={2024}
}

@article{aditya2025can,
  title={Can LLMs Reliably Simulate Real Students' Abilities in Mathematics and Reading Comprehension?},
  author={Aditya Srivatsa, KV and Maurya, Kaushal Kumar and Kochmar, Ekaterina},
  journal={arXiv e-prints},
  pages={arXiv--2507},
  year={2025}
}

@inproceedings{sauberli2025llms,
  title     = {Do {LLM}s Give Psychometrically Plausible Responses in Educational Assessments?},
  author    = {S{\"a}uberli, Andreas and Frassinelli, Diego and Plank, Barbara},
  booktitle = {Proceedings of the 20th Workshop on Innovative Use of NLP for Building Educational Applications ({BEA} 2025)},
  month     = jul,
  year      = {2025},
  address   = {Vienna, Austria},
  publisher = {Association for Computational Linguistics},
  pages     = {266--278},
  doi       = {10.18653/v1/2025.bea-1.21},
  url       = {https://aclanthology.org/2025.bea-1.21/}
}

@inproceedings{han-etal-2025-leveraging-fine,
    title = "Leveraging Fine-tuned Large Language Models in Item Parameter Prediction",
    author = "Han, Suhwa  and
      Rijmen, Frank  and
      Boykin, Allison Ames  and
      Lottridge, Susan",
    editor = "Wilson, Joshua  and
      Ormerod, Christopher  and
      Beiting Parrish, Magdalen",
    booktitle = "Proceedings of the Artificial Intelligence in Measurement and Education Conference (AIME-Con): Full Papers",
    month = oct,
    year = "2025",
    address = "Wyndham Grand Pittsburgh, Downtown, Pittsburgh, Pennsylvania, United States",
    publisher = "National Council on Measurement in Education (NCME)",
    url = "https://aclanthology.org/2025.aimecon-main.27/",
    pages = "250--264",
    ISBN = "979-8-218-84228-4"
}

@inproceedings{nguyen-etal-2025-qg,
    title = "{QG}-{SMS}: Enhancing Test Item Analysis via Student Modeling and Simulation",
    author = "Nguyen, Bang  and
      Du, Tingting  and
      Yu, Mengxia  and
      Angrave, Lawrence  and
      Jiang, Meng",
    editor = "Che, Wanxiang  and
      Nabende, Joyce  and
      Shutova, Ekaterina  and
      Pilehvar, Mohammad Taher",
    booktitle = "Proceedings of the 63rd Annual Meeting of the Association for Computational Linguistics (Volume 1: Long Papers)",
    month = jul,
    year = "2025",
    address = "Vienna, Austria",
    publisher = "Association for Computational Linguistics",
    url = "https://aclanthology.org/2025.acl-long.1268/",
    doi = "10.18653/v1/2025.acl-long.1268",
    pages = "26152--26168",
    ISBN = "979-8-89176-251-0"
}

@article{moses2017review,
  title={A review of developments and applications in item analysis},
  author={Moses, Tim},
  journal={Advancing human assessment: The methodological, psychological and policy contributions of ETS},
  pages={19--46},
  year={2017},
  publisher={Springer}
}

@book{crocker1986introduction,
  title={Introduction to classical and modern test theory.},
  author={Crocker, Linda and Algina, James},
  year={1986},
  publisher={ERIC}
}

@book{haladyna2013developing,
  title={Developing and validating test items},
  author={Haladyna, Thomas M and Rodriguez, Michael C},
  year={2013},
  publisher={Routledge}
}

@article{eignor2013standards,
  title={The standards for educational and psychological testing.},
  author={Eignor, Daniel R},
  year={2013},
  publisher={American Psychological Association}
}

@book{lord2008statistical,
  title={Statistical theories of mental test scores},
  author={Lord, Frederic M and Novick, Melvin R},
  year={2008},
  publisher={IAP}
}

@article{ebel1972essentials,
  title={Essentials of educational measurement},
  author={Ebel, Robert L and Frisbie, David A},
  year={1972},
  publisher={Prentice-Hall Englewood Cliffs, NJ}
}

@article{mccowan1999item,
  title={Item Analysis for Criterion-Referenced Tests.},
  author={McCowan, Richard J and McCowan, Sheila C},
  journal={Online Submission},
  year={1999},
  publisher={ERIC}
}

@article{alkhuzaey2024text,
  title={Text-based question difficulty prediction: A systematic review of automatic approaches},
  author={AlKhuzaey, Samah and Grasso, Floriana and Payne, Terry R and Tamma, Valentina},
  journal={International Journal of Artificial Intelligence in Education},
  volume={34},
  number={3},
  pages={862--914},
  year={2024},
  publisher={Springer}
}

@inproceedings{veeramani2024large,
  title={Large language model-based pipeline for item difficulty and response time estimation for educational assessments},
  author={Veeramani, Hariram and Thapa, Surendrabikram and Shankar, Natarajan Balaji and Alwan, Abeer},
  booktitle={Proceedings of the 19th Workshop on Innovative Use of NLP for Building Educational Applications (BEA 2024)},
  pages={561--566},
  year={2024}
}

@article{liusie2023analysis,
  title={Analysis of the cambridge multiple-choice questions reading dataset with a focus on candidate response distribution},
  author={Liusie, Adian and Raina, Vatsal and Mullooly, Andrew and Knill, Kate and Gales, Mark JF},
  journal={arXiv preprint arXiv:2306.13047},
  year={2023}
}

@article{maeda2025field,
  title={Field-testing multiple-choice questions with AI examinees: English grammar items},
  author={Maeda, Hotaka},
  journal={Educational and Psychological Measurement},
  volume={85},
  number={2},
  pages={221--244},
  year={2025},
  publisher={Sage Publications Sage CA: Los Angeles, CA}
}

@article{sonkar2024llm,
  title={Llm-based cognitive models of students with misconceptions},
  author={Sonkar, Shashank and Chen, Xinghe and Liu, Naiming and Baraniuk, Richard G and Sachan, Mrinmaya},
  journal={arXiv preprint arXiv:2410.12294},
  year={2024}
}

@article{ross2025learning,
  title={Learning to make mistakes: Modeling incorrect student thinking and key errors},
  author={Ross, Alexis and Andreas, Jacob},
  journal={arXiv preprint arXiv:2510.11502},
  year={2025}
}

@misc{anthropic2024claude35haiku,
author       = {{Anthropic}},
title        = {Model Card Addendum: Claude 3.5 Haiku and Upgraded Claude 3.5 Sonnet},
year         = {2024},
howpublished = {\url{https://assets.anthropic.com/m/1cd9d098ac3e6467/original/Claude-3-Model-Card-October-Addendum.pdf}},
note         = {Accessed: 2026-06-12}
}

@misc{meta2024llama32,
author       = {{Meta AI}},
title        = {Llama 3.2: Revolutionizing Edge AI and Vision with Open, Customizable Models},
year         = {2024},
howpublished = {\url{https://ai.meta.com/blog/llama-3-2-connect-2024-vision-edge-mobile-devices/}},
note         = {Accessed: 2026-06-12}
}

@misc{meta2024llama33,
  author       = {{Meta AI}},
  title        = {Meta Llama 3.3 70B Instruct Model Card},
  year         = {2024},
  howpublished = {\url{https://huggingface.co/meta-llama/Llama-3.3-70B-Instruct}},
  note         = {Accessed: 2026-06-12}
}

@misc{jiang2023mistral7b,
      title={Mistral 7B}, 
      author={Albert Q. Jiang and Alexandre Sablayrolles and Arthur Mensch and Chris Bamford and Devendra Singh Chaplot and Diego de las Casas and Florian Bressand and Gianna Lengyel and Guillaume Lample and Lucile Saulnier and Lélio Renard Lavaud and Marie-Anne Lachaux and Pierre Stock and Teven Le Scao and Thibaut Lavril and Thomas Wang and Timothée Lacroix and William El Sayed},
      year={2023},
      eprint={2310.06825},
      archivePrefix={arXiv},
      primaryClass={cs.CL},
      url={https://arxiv.org/abs/2310.06825}, 
}

@article{olmo20242,
  title={2 OLMo 2 Furious},
  author={OLMo, Team and Walsh, Pete and Soldaini, Luca and Groeneveld, Dirk and Lo, Kyle and Arora, Shane and Bhagia, Akshita and Gu, Yuling and Huang, Shengyi and Jordan, Matt and others},
  journal={arXiv preprint arXiv:2501.00656},
  year={2024}
}

@article{abouelenin2025phi,
  title={Phi-4-mini technical report: Compact yet powerful multimodal language models via mixture-of-loras},
  author={Abouelenin, Abdelrahman and Ashfaq, Atabak and Atkinson, Adam and Awadalla, Hany and Bach, Nguyen and Bao, Jianmin and Benhaim, Alon and Cai, Martin and Chaudhary, Vishrav and Chen, Congcong and others},
  journal={arXiv preprint arXiv:2503.01743},
  year={2025}
}

@article{yuan2026towards,
  title={Towards Valid Student Simulation with Large Language Models},
  author={Yuan, Zhihao and Xiao, Yunze and Li, Ming and Xuan, Weihao and Tong, Richard and Diab, Mona and Mitchell, Tom},
  journal={arXiv preprint arXiv:2601.05473},
  year={2026}
}

@article{peters2025text,
  title={Text-Based Approaches to Item Difficulty Modeling in Large-Scale Assessments: A Systematic Review},
  author={Peters, Sydney and Zhang, Nan and Jiao, Hong and Li, Ming and Zhou, Tianyi and Lissitz, Robert},
  journal={arXiv preprint arXiv:2509.23486},
  year={2025}
}

@article{wang2026cognitive,
  title={Cognitive Episodes in LLM Reasoning Traces Enable Interpretable Human Item Difficulty Prediction},
  author={Wang, Chenguang and Li, Ming and Zeng, Xinyue and Li, Zhuochun and Jiao, Hong and Zhou, Tianyi and Zhou, Dawei},
  journal={arXiv preprint arXiv:2606.28186},
  year={2026}
}

@misc{openai2026gpt55,
  author = {{OpenAI}},
  title = {{GPT}-5.5 System Card},
  year = {2026},
  month = apr,
  howpublished = {\url{https://openai.com/index/gpt-5-5-system-card/}},
  note = {Published April 23, 2026; updated April 24, 2026; accessed August 1, 2026}
}

@inproceedings{benedetto2024using,
  title={Using LLMs to simulate students’ responses to exam questions},
  author={Benedetto, Luca and Aradelli, Giovanni and Donvito, Antonia and Lucchetti, Alberto and Cappelli, Andrea and Buttery, Paula},
  booktitle={Findings of the Association for Computational Linguistics: EMNLP 2024},
  pages={11351--11368},
  year={2024}
}

\clearpage
\appendix
\section{Related Work}
\label{sec:related-work}

\subsection{Item Difficulty Prediction}

Item difficulty is central to educational assessment, informing test construction, adaptive testing, and learning personalization. Conventionally, it is estimated from examinee response data, as facility (proportion correct) under classical test theory (CTT) or as a model-based parameter under item response theory (IRT) and Rasch models~\citep{hambleton1991fundamentals, demars2010item, hsu2018automated}. Since field testing is costly and hard to scale, prior work has explored text-based difficulty prediction using various item-level features, as well as machine learning and deep learning models~\citep{sano2015automated, loukina2016textual, huang2017question, devlin2019bert, he2021automatically, benedetto2023quantitative}. More recent studies adopt LLMs to predict difficulty directly or to derive auxiliary signals such as reasoning traces and uncertainty estimates~\citep{rogoz2024unibucllm, zotos2024you, li2025item, feng2025reasoning}. However, difficulty alone offers only a partial view of item quality, since items that appear textually complex may not necessarily discriminate readers' level. Moreover, although LLM uncertainty has been explored as a signal for item difficulty, such model-centric uncertainty should not be assumed to reflect the difficulty experienced by human test-takers \citep{zotos2024you,li2025can}. We therefore argue that difficulty is important but insufficient, and examine whether LLMs can also predict item discrimination, i.e., how well items distinguish different student levels.

\subsection{Item Discrimination Prediction}

Psychometrically, item discrimination is operationalized through item-total correlations in CTT or the IRT slope parameter~\citep{demars2010item, lord2012applications, embretson2025item}. Unlike difficulty, which summarizes average challenge, discrimination requires modeling response variation across ability groups, making it harder to infer from item text alone and less studied in prior work. Recent work predicts both difficulty and discrimination from item text and structured attributes, finding discrimination harder to model and more dependent on explicit metadata~\citep{han-etal-2025-leveraging-fine}. LLM-based item analysis has also begun to evaluate discrimination through simulated student responses~\citep{nguyen-etal-2025-qg}. We therefore ask whether LLM-based methods can predict how item success covaries with human proficiency, rather than merely whether an item is hard.

\subsection{Synthetic Responses for Educational Assessment}

Student simulation supports tutoring, learning-by-teaching, and formative evaluation, traditionally relying on hand-crafted misconception models, rule-based dialogue policies, or human role-play data~\citep{vanlehn1994applications, chrysafiadi2013student, graesser2005autotutor, stasaski2020more}. LLMs make simulation more scalable through role prompting, fine-tuning,
and agentic designs that emulate diverse learner profiles, sometimes
combined with knowledge tracing or IRT-style ability representations
\citep{markel2023gpteach,lee2023generative,park2023generative,
kaser2024simulated,corbett1994knowledge,rasch1993probabilistic,
lord2012applications,embretson2025item,miroyan2025parastudent,
ross2025modeling}.

Proficiency-conditioned prompting has also been studied for simulating
students at different skill levels, although the resulting ability
patterns may not generalize consistently across models
\citep{benedetto2024using}.

For assessment, recent work uses heterogeneous or ability-aligned LLMs
to generate synthetic response matrices and estimate item parameters
through IRT-style calibration
\citep{park2024large,liu2025leveraging}.

Other work applies IRT-style models to compare LLM and human ability
patterns \citep{aditya2025can}. However, prior studies caution that fluent student-like responses or lower accuracy do not ensure behavioral fidelity, as LLMs may misalign with human error distributions, item facility, distractor choices, or ability-specific response patterns~\citep{hayakawa2024can, sauberli2025llms, aditya2025can}. Recent work further argues that valid LLM-based student simulation should prioritize epistemic fidelity over surface-level realism, since highly capable LLMs may exhibit a competence paradox when asked to emulate partially knowledgeable learners \citep{yuan2026towards}. We therefore evaluate synthetic LLM responses by alignment with human item discrimination, asking whether models succeed and fail in ways that preserve how items differentiate students by proficiency.

\section{Model Answer Accuracy}
\label{app:accuracy}

Table~\ref{tab:answer_accuracy} reports the answer accuracy of each model under the no-persona baseline and three proficiency-prompt conditions. Overall, stronger proprietary models achieve consistently high accuracy, with GPT-5 and GPT-5.5 reaching the highest average correct rates. In contrast, smaller open-weight models show substantially lower answer accuracy, suggesting that their synthetic response patterns may be constrained by limited reading-comprehension capability. Across most models, the low-, medium-, and high-proficiency prompts introduce only small changes in accuracy, indicating that proficiency prompting does not reliably induce large proficiency-separated performance gaps at the aggregate answer level.

\begin{table}[t]
\centering
\resizebox{\columnwidth}{!}{
\begin{tabular}{lccccc}
\toprule
\textbf{Model} & \textbf{Default} & \textbf{Low} & \textbf{Medium} & \textbf{High} & \textbf{Average} \\
\midrule
Llama 2-7B & 0.410 & 0.406 & 0.420 & 0.417 & 0.413 \\
Llama 2-13B & 0.482 & 0.448 & 0.464 & 0.450 & 0.461 \\
Llama 3.1-8B & 0.662 & 0.652 & 0.670 & 0.657 & 0.660 \\
Llama 3.2-1B & 0.212 & 0.173 & 0.192 & 0.183 & 0.190 \\
Llama 3.2-3B & 0.436 & 0.424 & 0.422 & 0.414 & 0.424 \\
Llama 3.3-70B & 0.813 & 0.817 & 0.832 & 0.810 & 0.818 \\
Mistral-7B-v0.3 & 0.415 & 0.380 & 0.393 & 0.392 & 0.395 \\
OLMo 2-7B & 0.445 & 0.435 & 0.425 & 0.449 & 0.439 \\
OLMo 2-13B & 0.533 & 0.598 & 0.609 & 0.591 & 0.583 \\
Phi-3-mini & 0.499 & 0.493 & 0.493 & 0.511 & 0.499 \\
Phi-3.5-mini & 0.566 & 0.573 & 0.545 & 0.557 & 0.560 \\
Phi-4-mini & 0.459 & 0.489 & 0.458 & 0.470 & 0.469 \\
Phi-4 & 0.810 & 0.820 & 0.812 & 0.806 & 0.812 \\
\midrule
Qwen2.5-0.5B & 0.141 & 0.110 & 0.112 & 0.105 & 0.117 \\
Qwen2.5-1.5B & 0.141 & 0.221 & 0.235 & 0.257 & 0.213 \\
Qwen2.5-3B & 0.591 & 0.623 & 0.629 & 0.623 & 0.617 \\
Qwen2.5-7B & 0.689 & 0.697 & 0.676 & 0.705 & 0.692 \\
Qwen2.5-14B & 0.739 & 0.765 & 0.760 & 0.767 & 0.758 \\
Qwen2.5-32B & 0.792 & 0.815 & 0.815 & 0.808 & 0.807 \\
Qwen3-0.6B & 0.475 & 0.450 & 0.469 & 0.478 & 0.468 \\
Qwen3-1.7B & 0.622 & 0.624 & 0.627 & 0.620 & 0.623 \\
Qwen3-4B & 0.755 & 0.758 & 0.760 & 0.745 & 0.755 \\
Qwen3-8B & 0.772 & 0.757 & 0.779 & 0.758 & 0.766 \\
Qwen3-14B & 0.861 & 0.860 & 0.871 & 0.865 & 0.864 \\
Qwen3-32B & 0.875 & 0.875 & 0.883 & 0.863 & 0.874 \\
Qwen3-235B-A22B & 0.910 & 0.898 & 0.904 & 0.912 & 0.906 \\
Qwen3.5-9B & 0.822 & 0.837 & 0.836 & 0.834 & 0.832 \\
Qwen3.5-27B & 0.908 & 0.907 & 0.917 & 0.907 & 0.910 \\
Qwen3.5-122B-A10B & 0.904 & 0.865 & 0.888 & 0.917 & 0.893 \\
\midrule
Claude 3.5 Haiku & 0.759 & 0.763 & 0.753 & 0.774 & 0.762 \\
Claude 3.7 Sonnet & 0.932 & 0.909 & 0.926 & 0.919 & 0.922 \\
Gemini 2.0 Flash & 0.868 & 0.881 & 0.889 & 0.893 & 0.883 \\
Gemini 2.5 Flash & 0.913 & 0.912 & 0.910 & 0.914 & 0.912 \\
Gemini 2.5 Pro & 0.918 & 0.832 & 0.859 & 0.889 & 0.875 \\
GPT-3.5-Turbo & 0.643 & 0.646 & 0.622 & 0.631 & 0.635 \\
GPT-4.1-mini & 0.897 & 0.870 & 0.897 & 0.880 & 0.886 \\
GPT-4.1 & 0.921 & 0.924 & 0.922 & \underline{0.922} & 0.922 \\
GPT-4o-mini & 0.793 & 0.770 & 0.786 & 0.797 & 0.787 \\
GPT-4o & 0.902 & 0.893 & 0.899 & 0.899 & 0.898 \\
GPT-o4-mini & 0.914 & 0.898 & 0.900 & 0.907 & 0.905 \\
GPT-5 & \underline{0.961} & \underline{0.960} & \underline{0.958} & \textbf{0.963} & \underline{0.961} \\
GPT-5.5 & \textbf{0.962} & \textbf{0.961} & \textbf{0.960} & \textbf{0.963} & \textbf{0.962} \\
\bottomrule
\end{tabular}
}
\vspace{-2.2mm}
\caption{Model answer accuracy under different proficiency-prompt conditions. Default denotes the no-persona setting, Low/Medium/High denote low-, medium-, and high-proficiency prompts, and Average reports the mean accuracy across these four settings. Proficiency prompting produces only small aggregate accuracy changes for most models.}
\label{tab:answer_accuracy}
\vspace{-3.2mm}
\end{table}

\section{Prompt Templates}
\label{app:prompts}

We used two prompt templates: one for direct discrimination prediction and one for response-based proxy estimation. For prompted-profile runs, the corresponding system message was prepended to the user prompt. For the no-persona setting, no system message was used. Figure~\ref{fig:direct-discrimination-prompt} shows the prompt used for direct discrimination prediction, where the gold answer was provided. Figure~\ref{fig:response-based-answer-prompt} shows the answer-collection prompt used for response-based proxy estimation, where the gold answer was withheld.

\begin{figure*}[t]
\centering
\begin{tcolorbox}[promptbox, title={Prompt Template for Direct Discrimination Prediction}]
\begin{Verbatim}[fontsize=\footnotesize, breaklines=true, breakanywhere=false, breaksymbolleft={}, breakindent=0pt]
(System Prompt)
No persona: [empty]

Low-proficiency persona:
Suppose you are a student taking the Cambridge English Test. You are a weak student with low-level English proficiency.

Medium persona:
Suppose you are a student taking the Cambridge English Test. You are an average student with medium-level English proficiency.

High-proficiency persona:
Suppose you are a student taking the Cambridge English Test. You are a good student with high-level English proficiency.

(User Prompt)
Analyze the discrimination value of the question. The discrimination value is the point-biserial correlation between selecting the correct answer and the test-taker's total test score, ranging from -1 to 1. A value near 0 means the item cannot distinguish between proficiency levels; a positive value means higher-proficiency students are more likely to answer correctly; a negative value means the item discriminates in the wrong direction. Predict the discrimination value and provide the final value in \boxed{...}:

Below is a Multiple Choice Question for Reading Comprehension.
Question: {QUESTION}
Options:
(A) {OPTION_A}
(B) {OPTION_B}
(C) {OPTION_C}
(D) {OPTION_D}
Reference Passage: {REFERENCE_PASSAGE}
Common European Framework of Reference for Languages (CEFR) Level: {CEFR_LEVEL}
Correct answer: {CORRECT_ANSWER}
\end{Verbatim}
\end{tcolorbox}
\caption{Prompt template for direct discrimination prediction. The gold answer was provided only in this setting.}
\label{fig:direct-discrimination-prompt}
\end{figure*}

\begin{figure*}[t]
\centering
\begin{tcolorbox}[promptbox, title={Answer Collection for Response-Based Proxy Estimation}]
\begin{Verbatim}[fontsize=\footnotesize, breaklines=true, breakanywhere=false, breaksymbolleft={}, breakindent=0pt]
(System Prompt)
No persona: [empty]

Low-proficiency persona:
Suppose you are a student taking the Cambridge English Test. You are a weak student with low-level English proficiency.

Medium persona:
Suppose you are a student taking the Cambridge English Test. You are an average student with medium-level English proficiency.

High-proficiency persona:
Suppose you are a student taking the Cambridge English Test. You are a good student with high-level English proficiency.

(User Prompt)
Answer the question below step by step, and provide the final answer in \boxed{...}:

Below is a Multiple Choice Question for Reading Comprehension.
Question: {QUESTION}
Options:
(A) {OPTION_A}
(B) {OPTION_B}
(C) {OPTION_C}
(D) {OPTION_D}
Reference Passage: {REFERENCE_PASSAGE}
Common European Framework of Reference for Languages (CEFR) Level: {CEFR_LEVEL}
\end{Verbatim}
\end{tcolorbox}
\caption{Prompt template for response-based proxy estimation. The correct answer was not included in the prompt.}
\label{fig:response-based-answer-prompt}
\end{figure*}

\section{Details for Diagnostic Experiments}
\label{app:revision_experiments}

This appendix provides complete protocols and supplementary results for the population, few-shot, supervised, respondent-validity, and error analyses summarized in the main text.

\subsection{Population-Aware Evaluation}
\label{app:population_aware}

\paragraph{Population structure.} Item discrimination is conditional on the examinee population used for calibration. The Cambridge items span four CEFR levels, and the human discrimination distribution differs across these strata (Table~\ref{tab:cefr_breakdown}). In particular, mean discrimination decreases from 0.414 at B1 to 0.239 at C2. An overall correlation can therefore combine within-level ranking signal with between-level differences.

\begin{table}[t]
\centering
\small
\resizebox{\linewidth}{!}{
\begin{tabular}{lrrrrrr}
\toprule
CEFR & Items & Tasks & Scaled Rasch diff. & Disc. mean & Disc. SD & Disc. $\leq .2$ \\
\midrule
B1 & 140 & 28 & 60.95 & 0.414 & 0.134 & 10 \\
B2 & 422 & 58 & 64.33 & 0.336 & 0.128 & 47 \\
C1 & 169 & 25 & 73.18 & 0.304 & 0.105 & 23 \\
C2 &  62 &  9 & 78.15 & 0.239 & 0.134 & 23 \\
\midrule
All & 793 & 120 & 66.70 & 0.335 & 0.133 & 103 \\
\bottomrule
\end{tabular}}
\caption{Dataset composition by CEFR level. Difficulty is the released scaled Rasch value, not proportion correct. Disc.\ denotes human-calibrated item discrimination.}
\label{tab:cefr_breakdown}
\end{table}

\paragraph{Metrics.} In addition to overall rank correlation, we report separate Spearman correlations for B1, B2, C1, and C2 and their unweighted macro average as diagnostics of level-specific stability. For the primary within-CEFR statistic, we rank predictions and labels separately within each CEFR level, standardize each within-level rank vector, and pool the standardized ranks. This equals an item-weighted mean of the level-specific correlations and makes no cross-level rank comparisons. Confidence intervals use 1,000 bootstrap samples with the 120 reading tasks as clusters.

\paragraph{Direct discrimination prediction.} Population-aware evaluation changes the interpretation of the negative overall correlations but not the main conclusion. For most models, negative overall correlations become approximately zero when only within-CEFR ranks are compared, indicating little reliable within-population ranking signal rather than systematic reverse ranking. GPT-5.5 is the main exception: under the low-proficiency prompt, its overall correlation is 0.152 and its within-CEFR correlation is 0.172 (95\% CI [0.109, 0.242]). Even this selected result remains weak as a prediction of human item discrimination.

\begin{figure*}[t]
\centering
\includegraphics[width=0.92\textwidth]{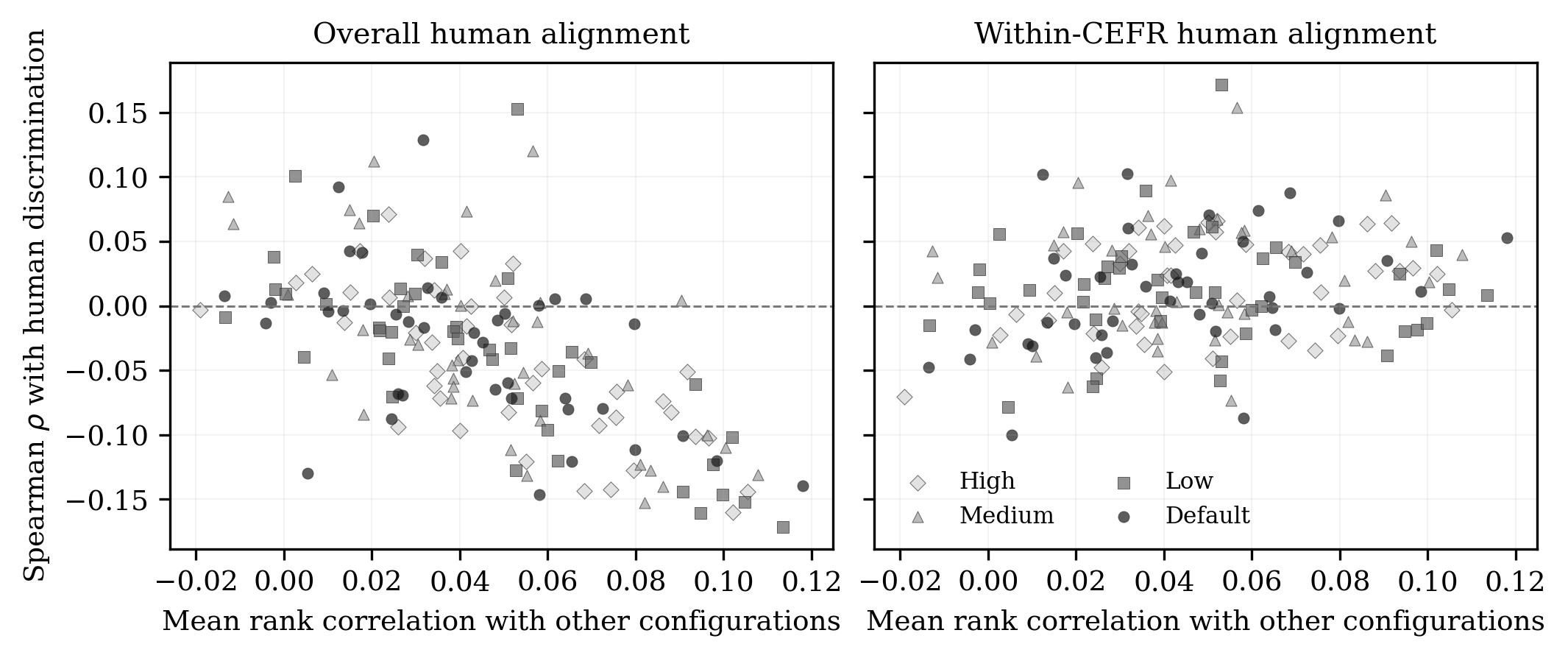}
\caption{Population adjustment of the direct-prediction diagnostic. Each point is one model-prompt configuration; shades and markers denote prompt conditions. Human alignment is overall on the left and uses standardized within-CEFR ranks on the right. Inter-model agreement is weak in both panels, and the negative association among comparatively higher-agreement configurations in the overall evaluation largely collapses within CEFR.}
\label{fig:consensus_population_adjustment}
\end{figure*}

Because this configuration is selected from 168 alternatives, its ordinary bootstrap interval is descriptive rather than an out-of-sample model-selection estimate.

We also test whether including population information in the prompt is sufficient. For GPT-4.1, removing the CEFR label changes overall correlation from $-0.140$ to 0.005, suggesting that the model sometimes treats a higher CEFR label as evidence of higher discrimination. A richer textual description of the target population does not repair the overall error ($\rho=-0.097$). For GPT-5.5, removing the CEFR label changes within-CEFR correlation only from 0.103 to 0.092, while the richer population description reduces it to 0.033. Coarse population descriptions are therefore neither necessary for the strongest model nor a reliable remedy for weaker models.

\paragraph{Response-based proxy estimation.} For each target item, we compute response-vector scores using only items from the same CEFR level. The primary estimator is the corrected item-rest correlation, $T_{m,p,-i}^{(s,\ell)}=\sum_{j:c_j=\ell,j\neq i}v_{j,m,p}$, which avoids part-whole inflation. For the full heterogeneous pool, it gives overall $\rho=0.255$ (95\% CI [0.194, 0.323]), within-CEFR $\rho=0.231$ (95\% CI [0.168, 0.300]), and RMSE 0.270.

The uncorrected same-CEFR item-total version gives overall $\rho=0.249$, within-CEFR $\rho=0.230$, and RMSE 0.275. We retain it as a comparability analysis because the released description uses a point-biserial item-total definition, but the underlying human response matrix and exact scoring implementation are unavailable. The close results show that focal-item correction does not create the positive correlation obtained by response-based proxy estimation. The original human test forms remain unavailable. Both estimators use a CEFR-wide synthetic capability score and are evaluated only by empirical alignment with the released human target.

Table~\ref{tab:ctt_null} reports the accuracy-conditioned null comparison.
\begin{table}[t]
\centering
\small
\resizebox{\linewidth}{!}{
\begin{tabular}{lrrrr}
\toprule
Metric & Observed & Null mean & Null 97.5\% & MC $p$ \\
\midrule
Within-CEFR $\rho$ & 0.231 & 0.095 & 0.155 & 0.001 \\
\bottomrule
\end{tabular}
}
\caption{Primary corrected item-rest proxy evaluated against 1,000 accuracy-conditioned permutations. Human discrimination is shuffled only within each CEFR level and synthetic-accuracy decile, preserving its coarse relationship with synthetic item accuracy. Monte Carlo $p$ uses the plus-one correction.}
\label{tab:ctt_null}
\end{table}

\begin{table}[t]
\centering
\small
\resizebox{\linewidth}{!}{
\begin{tabular}{lrrrr}
\toprule
Rest-pool length & Draws & Overall $\rho$ & Within-CEFR $\rho$ & Within-CEFR 95\% range \\
\midrule
20 & 200 & 0.251 & 0.227 & [0.212, 0.245] \\
40 & 200 & 0.252 & 0.229 & [0.217, 0.241] \\
80 & 200 & 0.253 & 0.230 & [0.223, 0.237] \\
\bottomrule
\end{tabular}}
\caption{Sensitivity to synthetic rest-pool length. For each item and draw, rest items are sampled without replacement from the same CEFR stratum. Values are means across draws. The final column is the 2.5th to 97.5th percentile range across subset draws rather than a sampling CI. C2 contains only 61 available rest items, so the nominal 80-item condition uses all 61.}
\label{tab:ctt_pool_sensitivity}
\end{table}

The accuracy-conditioned null preserves the coarse relationship between the human target and synthetic item facility: within each CEFR level, items are divided into deciles by full-pool synthetic accuracy, and human discrimination labels are permuted only within these cells. Across 1,000 permutations, mean within-CEFR correlation is 0.095 and the 97.5th percentile is 0.155, below the observed 0.231 ($p=0.001$ with plus-one correction). The proxy therefore contains ranking information beyond coarse synthetic accuracy, although this comparison does not establish that the synthetic pool matches the human calibration population.

Table~\ref{tab:ctt_protocol_comparison} summarizes the global-score, CEFR-stratified, and focal-item-corrected protocols.
\begin{table}[t]
\centering
\small
\resizebox{\linewidth}{!}{
\begin{tabular}{lrrrrr}
\toprule
Pool & Global item-total & CEFR item-total & CEFR item-rest & Rest within-CEFR & Rest RMSE \\
\midrule
Default & 0.204 & 0.212 & 0.216 & 0.210 & 0.284 \\
Low & 0.231 & 0.237 & 0.242 & 0.218 & 0.281 \\
Medium & 0.196 & 0.206 & 0.211 & 0.187 & 0.286 \\
High & 0.234 & 0.238 & 0.243 & 0.214 & 0.280 \\
Full pool & 0.243 & 0.249 & 0.255 & 0.231 & 0.270 \\
\bottomrule
\end{tabular}}
\caption{Uncorrected global and same-CEFR item-total estimates compared with the primary corrected same-CEFR item-rest estimator, recomputed from the 42-model response matrix using the answer-parsing rules described in Appendix~\ref{app:model_details}. Default denotes the no-persona condition.}
\label{tab:ctt_protocol_comparison}
\end{table}

\subsection{Few-Shot and Supervised Comparisons}
\label{app:fewshot_baselines}

\paragraph{Few-shot design.} We evaluate GPT-5.5, GPT-4.1, and Qwen3-32B with four demonstrations sampled as three fixed sets. Demonstrations are task-disjoint from the target item, matched by CEFR level, and stratified across the observed discrimination range. We additionally evaluate one eight-shot run for GPT-5.5 and Qwen3-32B. Table~\ref{tab:fewshot_results} reports zero-shot results, each four-shot run, and the mean of its three evaluation metrics.

\begin{table*}[t]
\centering
\small
\begin{tabular}{llrrr}
\toprule
Model & Setting & Overall $\rho$ & Within-CEFR $\rho$ & RMSE \\
\midrule
GPT-5.5 & Zero-shot & 0.129 & 0.103 & 0.154 \\
        & Four-shot, set 1 & 0.278 & 0.043 & 0.140 \\
        & Four-shot, set 2 & 0.239 & 0.015 & 0.141 \\
        & Four-shot, set 3 & 0.273 & 0.055 & 0.141 \\
        & Four-shot mean & 0.263 & 0.038 & 0.141 \\
        & Eight-shot & 0.261 & 0.013 & 0.138 \\
\midrule
GPT-4.1 & Zero-shot & $-0.140$ & 0.053 & 0.193 \\
        & Four-shot, set 1 & 0.192 & 0.021 & 0.142 \\
        & Four-shot, set 2 & 0.161 & $-0.032$ & 0.146 \\
        & Four-shot, set 3 & 0.174 & $-0.024$ & 0.143 \\
        & Four-shot mean & 0.176 & $-0.012$ & 0.144 \\
\midrule
Qwen3-32B & Zero-shot & $-0.043$ & 0.025 & 0.211 \\
           & Four-shot, set 1 & 0.237 & 0.024 & 0.139 \\
           & Four-shot, set 2 & 0.229 & 0.016 & 0.139 \\
           & Four-shot, set 3 & 0.225 & $-0.010$ & 0.144 \\
           & Four-shot mean & 0.231 & 0.010 & 0.140 \\
           & Eight-shot & 0.296 & 0.057 & 0.142 \\
\bottomrule
\end{tabular}
\caption{Few-shot direct discrimination prediction. Zero-shot and few-shot rows use the no-persona condition with identical target items and CEFR metadata. Demonstrations improve overall correlation and RMSE but do not consistently improve within-CEFR ranking. ``Four-shot mean'' averages the three run-level metrics.}
\label{tab:fewshot_results}
\end{table*}

All nine individual four-shot runs improve overall correlation relative to their paired zero-shot run. However, no model improves its mean within-CEFR correlation across the three demonstration sets. The demonstrations also compress 98 to 100\% of predictions into the $[0.2,0.6]$ interval. For GPT-5.5, increasing from zero to eight shots reduces within-CEFR correlation from 0.103 to 0.013. Qwen3-32B's eight-shot run increases within-CEFR correlation numerically from 0.025 to 0.057, but remains weak.

The few-shot gain is therefore more consistent with learning the CEFR-level numerical range and shrinking predictions toward typical values than with improving item ranking within the target population.

\paragraph{Supervised comparisons.} We construct five-fold out-of-fold predictions using splits that are grouped by reading task and stratified by CEFR level. The deployable text-only comparisons are TF-IDF ridge regression, ridge regression over frozen \texttt{bge-large-en-v1.5} embeddings, and fine-tuned \texttt{answerdotai/ModernBERT-base}. The encoder result averages three training seeds. We also report a CEFR-mean prior and a ridge model using scaled human Rasch difficulty. The latter requires pretest information and is an explanatory control rather than a deployable text-only method.

All reported text-only models serialize the question, four options, correct-answer marker, and passage. CEFR is used only to stratify the fixed task-grouped folds and to compute evaluation metrics. It is not included as an input feature in the reported rows. TF-IDF uses word unigrams and bigrams (minimum document frequency 2, at most 50,000 features) with fold-internal ridge selection over 13 log-spaced penalties from $10^{-2}$ to $10^4$. Frozen embeddings are produced by \texttt{BAAI/bge-large-en-v1.5} and fit with the same ridge protocol.

ModernBERT is fine-tuned as a scalar regressor with mean-squared error, maximum sequence length 1,024, batch size 16, four epochs, learning rate $5\times10^{-5}$, AdamW weight decay 0.01, OneCycle scheduling with a 0.1 warm-up fraction, and gradient clipping at 1.0. Training uses bfloat16 and seeds 0, 1, and 2. Reported predictions average the three out-of-fold vectors. We did not use early stopping or a separate hyperparameter search. Long inputs are truncated by the model tokenizer after the serialized question and options, so truncation affects the passage tail.

Table~\ref{tab:supervised_results} reports the complete supervised comparison.
\begin{table}[t]
\centering
\small
\resizebox{\linewidth}{!}{
\begin{tabular}{lrr}
\toprule
Method & Overall $\rho$ & Within-CEFR $\rho$ \\
\midrule
GPT-5.5 Low (best within-CEFR) & 0.152 & 0.172 \\
CEFR-mean prior & 0.266 & \textemdash \\
TF-IDF ridge & 0.215 & $-0.039$ \\
Frozen embedding ridge & 0.158 & $-0.076$ \\
Fine-tuned encoder & 0.215 & 0.014 \\
\midrule
Rasch difficulty $\rightarrow$ discrimination ridge & 0.370 & 0.246 \\
CEFR item-rest, full pool & 0.255 & 0.231 \\
\bottomrule
\end{tabular}}
\caption{Comparison with supervised and non-LLM methods. All supervised results are out-of-fold. The Rasch-difficulty benchmark uses human pretest information and is not deployable without human responses.}
\label{tab:supervised_results}
\end{table}

The CEFR-mean prior attains a relatively high overall correlation because mean discrimination differs across CEFR levels. Conceptually, however, it assigns the same value to every item in a level and therefore provides no within-level ranking; its within-CEFR entry is consequently left undefined. Fold-specific training means can create incidental within-level variation in the out-of-fold prediction vector, but that variation is an artifact of the cross-validation partition rather than a property of the prior. Across the tested supervised text-only methods, within-CEFR correlations remain near zero. This does not prove that item text contains no discrimination information. Instead, these methods find no reliable within-CEFR information on this dataset. The Rasch-difficulty benchmark shows that a human response statistic does contain information about discrimination. The CEFR-stratified item-rest proxy similarly provides a positive within-CEFR correlation without using human-calibrated discrimination labels or other human response statistics, although it requires the answer key.

\begin{table*}[t]
\centering
\small
\setlength{\tabcolsep}{4pt}
\begin{tabular}{llrrrrrr}
\toprule
Type & Method & B1 $\rho$ & B2 $\rho$ & C1 $\rho$ & C2 $\rho$ & Macro $\rho$ & Within-CEFR $\rho$ \\
\midrule
Direct & GPT-5.5 Default & 0.048 & 0.120 & 0.070 & 0.195 & 0.108 & 0.103 \\
Direct & GPT-5.5 Low & 0.162 & 0.145 & 0.189 & 0.330 & 0.206 & 0.172 \\
Direct & All-models Default & 0.025 & 0.053 & $-0.045$ & $-0.065$ & $-0.008$ & 0.018 \\
\midrule
Response & Item-total, no-persona & 0.271 & 0.219 & 0.138 & 0.197 & 0.206 & 0.209 \\
Response & Item-total, full pool & 0.339 & 0.202 & 0.211 & 0.228 & 0.245 & 0.230 \\
\midrule
Prior & CEFR mean & \textemdash & \textemdash & \textemdash & \textemdash & \textemdash & \textemdash \\
Supervised & TF-IDF ridge & 0.099 & $-0.131$ & 0.057 & 0.016 & 0.010 & $-0.039$ \\
Supervised & Frozen embedding ridge & $-0.146$ & $-0.129$ & 0.021 & 0.181 & $-0.018$ & $-0.076$ \\
Supervised & ModernBERT & 0.013 & $-0.022$ & 0.153 & $-0.125$ & 0.005 & 0.014 \\
Human pretest & Rasch difficulty $\rightarrow$ discrimination & 0.331 & 0.226 & 0.261 & 0.145 & 0.241 & 0.246 \\
\bottomrule
\end{tabular}
\caption{Within-CEFR results for representative direct configurations, the uncorrected CEFR-stratified item-total comparator, and diagnostic controls. Macro $\rho$ is the unweighted mean of the four level-specific correlations. Within-CEFR $\rho$ ranks and standardizes within each level before pooling and is therefore item-weighted. The C2 stratum is smallest (62 items from 9 tasks), so its estimates are correspondingly less stable. The CEFR-mean prior assigns no item-specific ranking within a level; incidental differences between fold-specific training means are not evaluated as within-level signal.}
\label{tab:within_cefr_detail}
\end{table*}

\subsection{Validity of Prompted Respondent Variation}
\label{app:persona_validity}

We test whether the three generic persona prompts, with temperature-zero decoding and the fixed step-by-step answer prompt, induce the ordered ability pattern expected from human-aligned proficiency variation. Only 8 of 42 models satisfy $\mathrm{Acc}_{\mathrm{low}} < \mathrm{Acc}_{\mathrm{mid}} < \mathrm{Acc}_{\mathrm{high}}$. Averaged across models, the high versus low accuracy difference is only 0.56 percentage points. A GEE model fitted to item-level correctness estimates a high-versus-low log-odds difference of 0.034 (95\% CI [0.009, 0.060]). The effect is statistically detectable but too small to produce well-separated synthetic proficiency groups.

The median Jaccard overlap between low- and high-persona error sets is 0.569, and the median rate at which personas disagree on an answer is 0.230. More importantly, item-level persona sensitivity is not associated with human item discrimination ($\rho=-0.039$, 95\% CI [$-0.113$, 0.039]). Persona prompting therefore changes some answers without aligning with the ability-conditioned item pattern measured in human pretesting.

Pool ablations support the same conclusion under the corrected within-CEFR protocol. Treating the four prompt conditions of one model as a respondent pool yields mean $\rho=-0.006$ (range $-0.116$ to 0.058). Randomly retaining one prompt condition per model yields mean $\rho=0.211$ (95\% draw range [0.175, 0.242]). Equalizing the six model-family groups at eight respondents each yields mean $\rho=0.189$, compared with 0.204 for random size-matched pools; leave-one-family-out estimates range from 0.169 to 0.253. Thus, the result is not created by one family, but its magnitude depends moderately on pool composition. The positive correlation comes primarily from differences across models rather than ordered prompt differences within a model. These results are specific to deterministic generic role prompting and do not cover richer learner-state constraints or repeated stochastic response generation.

\subsection{Error and Distractor Analyses}
\label{app:error_distractor}

\paragraph{Shared heuristics and task robustness.} Among the 19 models with negative overall direct-prediction correlations, predictions correlate with passage length at 0.340 and with scaled human Rasch difficulty at 0.180 on average. For the two positively aligned models, the corresponding averages are 0.028 and 0.000. Scaled human Rasch difficulty is itself negatively correlated with discrimination ($\rho=-0.374$). These relationships explain why multiple models can agree with one another while being negatively aligned with overall human discrimination: they share difficulty- and length-related heuristics that interact with the dataset's CEFR structure.

The central results are not driven by a small number of reading tasks: leave-one-task-out changes in Spearman correlation are below 0.01 in absolute value. We additionally fit a task-random-intercept linear model whose dependent variable is human discrimination and whose focal fixed effect is either the GPT-5.5 direct prediction or the uncorrected CEFR-stratified item-total proxy. This regression uses the uncorrected response-based comparator from the audited diagnostic protocol rather than the primary corrected item-rest proxy. CEFR indicators, scaled human Rasch difficulty, synthetic accuracy, passage length, question length, option length, lexical diversity, readability, and negation indicators enter as controls. The standardized focal coefficients remain positive for GPT-5.5 direct prediction ($\beta=0.125$, $p<.001$) and the CEFR-stratified item-total proxy ($\beta=0.239$, $p<.001$). Synthetic accuracy is defined identically in both regressions as the fraction of all 168 model-prompt respondents that answer item $i$ correctly. These observational regressions are robustness checks rather than evidence of independent psychometric validity.

\paragraph{Distractors.} For 502 items with option-level human statistics, we compare synthetic option choices with human distractor behavior. Synthetic distractor selection rates correlate with human endorsement rates ($\rho=0.315$, task-clustered 95\% CI [0.252, 0.369]), the mean within-item distractor-ranking correlation is 0.269 (95\% CI [0.204, 0.332]), and the most attractive human distractor is identified on 53.9\% of items (95\% CI [49.2\%, 59.0\%]; tie-aware random-choice rate: 34.9\%). Of the 502 items with option-level statistics, 501 have sufficient nonconstant human endorsement values for within-item metrics. We compute Spearman correlation over the three distractors using average ranks for ties, omit undefined item-level correlations, and average the remaining item correlations. For top-distractor accuracy, a prediction is correct if the synthetic and human top-distractor sets overlap; this counts any distractor tied for highest human endorsement as correct. Human top ties occur in 24 of the 501 evaluable items. Thus, LLM responses contain some information about which distractors are attractive, although the top-choice result is the more directly interpretable statistic.

This signal does not extend to ability-conditioned distractor discrimination. Synthetic and human distractor discrimination correlate at only 0.086 (95\% CI [0.032, 0.136]), and the prompt-profile selection gap is uncorrelated with human distractor discrimination ($\rho=-0.013$). Synthetic choice entropy is slightly higher for low-discrimination items (1.016 for human discrimination $\leq0.2$, versus 0.901 otherwise), suggesting a weak screening cue but not a substitute for option-level human calibration.

\subsection{Model and Inference Details}
\label{app:model_details}

The proprietary subset includes GPT-3.5-Turbo~\citep{openai_chatgpt_whisper_api_2023}, GPT-4o-mini~\citep{openai_gpt4o_mini_2024}, GPT-4o~\citep{hurst2024gpt}, GPT-4.1-mini and GPT-4.1~\citep{openai_gpt4_1_2025}, GPT-o4-mini~\citep{openai2025o3o4mini}, GPT-5~\citep{openai_gpt5_system_card_2025}, and GPT-5.5~\citep{openai2026gpt55}; Claude 3.5 Haiku~\citep{anthropic2024claude35haiku} and Claude 3.7 Sonnet~\citep{anthropic2025claude37sonnet}; and Gemini 2.0 Flash~\citep{google2024gemini20}, Gemini 2.5 Flash, and Gemini 2.5 Pro~\citep{comanici2025gemini}.

The open-weight subset includes Llama 2-7B and Llama 2-13B~\citep{touvron2023llama}, Llama 3.1-8B~\citep{grattafiori2024llama}, Llama 3.2-1B and Llama 3.2-3B~\citep{meta2024llama32}, and Llama 3.3-70B~\citep{meta2024llama33}; Mistral-7B-v0.3~\citep{jiang2023mistral7b}; OLMo 2-7B and OLMo 2-13B~\citep{olmo20242}; Phi-3-mini and Phi-3.5-mini~\citep{abdin2024phi3technicalreporthighly}, Phi-4~\citep{abdin2024phi}, and Phi-4-mini~\citep{abouelenin2025phi}; and Qwen2.5, Qwen3, and Qwen3.5 models at multiple scales~\citep{qwen2024qwen25,yang2025qwen3,qwen2026qwen35}. Model names and parameter scales are listed in Appendix~\ref{app:complete_results}.

Hosted models were invoked using routed model names (e.g., \texttt{openai/gpt-4.1}, \texttt{openai/gpt-5.5}, \texttt{anthropic/claude-3.7-sonnet}, and \texttt{google/gemini-2.5-pro}). Open-weight models use the corresponding model families and sizes with greedy decoding and half-precision weights when supported. API decoding used temperature 0 and a maximum of 4,096 output tokens. Task-specific parsers extract either a scalar prediction or an answer choice; response parsing includes a single-letter fallback. Discrimination predictions outside $[-1,1]$ are treated as invalid, and undefined item-rest or item-total correlations in response-based proxy estimation are assigned zero.

\section{Complete Results}
\label{app:complete_results}

\begin{figure*}[!t]
\centering
\includegraphics[width=\textwidth]{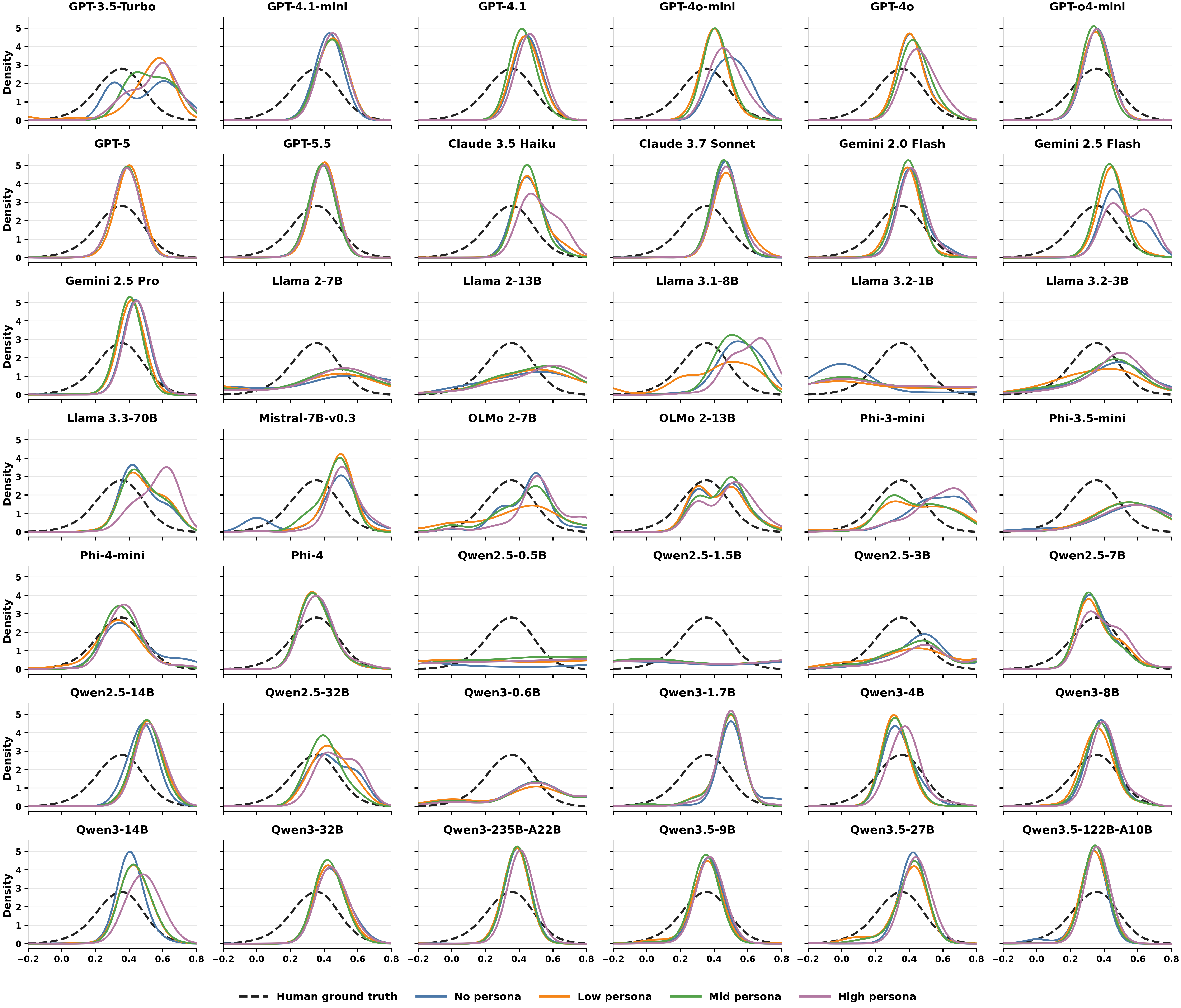}
\vspace{-1.5mm}
\caption{Complete distributions of human-calibrated and LLM-predicted item discrimination values for all 42 models. The x-axis shows the item discrimination value. The y-axis shows the estimated density of items at each discrimination value, where higher density indicates that more items fall within that range.}
\label{fig:direct_dist_all_42}
\vspace{-2.5mm}
\end{figure*}

This appendix provides complete model-level results for all 42 evaluated LLMs. Table~\ref{tab:table1_all} reports overall correlation and RMSE for every zero-shot model-prompt configuration, while Table~\ref{tab:direct_population} provides within-CEFR correlations and task-clustered confidence intervals for representative configurations. Figure~\ref{fig:direct_dist_all_42} shows the direct-prediction distributions for all 42 models. Table~\ref{tab:table3_all} reports high- and low-discrimination retrieval for the no-persona direct predictions and all five response-based pools. Its response-based rows use the audited uncorrected item-total comparator. For the full heterogeneous pool, focal-item correction leaves all four retrieval values unchanged. Table~\ref{tab:retrieval_uncertainty} reports clustered uncertainty and threshold-based screening for the primary corrected item-rest proxy. For predicted and gold sets of equal size, we define
\[
\mathrm{Overlap}@k =
\frac{
|\mathcal{S}^{k}_{\mathrm{gold}} \cap \mathcal{S}^{k}_{\mathrm{pred}}|
}{
|\mathcal{S}^{k}_{\mathrm{gold}}|
},
\]
which is equivalent to precision and recall at the cutoff. Random selection has expected overlap 0.10 at $k=10$ and 0.20 at $k=20$.

\subsection{Distributional Analysis Across Prompt Conditions}

Most models produce substantially narrower distributions than the human-calibrated distribution. Human item discrimination spans a broad range, reflecting item-level variation in how well questions distinguish higher- and lower-proficiency test-takers. In contrast, many LLMs concentrate their predictions within a small positive interval. In particular, LLMs rarely assign very low or negative discrimination values, placing little probability mass in the lower tail where weakly discriminating or potentially problematic items appear.

Proficiency-prompt conditions change the scale or shape of direct predictions for some models, but do not match the human-calibrated marginal distribution. For many models, the no-persona, low-, medium-, and high-proficiency curves largely overlap. When differences appear, they mainly involve location or dispersion shifts and are not directionally consistent across models.

Overall, these distributional results help explain the weak correlations: direct predictions do not merely differ in scale from human-calibrated discrimination, but also fail to preserve the item-level variation needed for reliable discrimination analysis.

\begin{table}[t]
\centering
\small
\resizebox{\linewidth}{!}{
\begin{tabular}{llrrrr}
\toprule
Model & Condition & Overall $\rho$ & 95\% CI & Within-CEFR $\rho$ & 95\% CI \\
\midrule
GPT-5.5 & Default & 0.129 & [0.058, 0.198] & 0.103 & [0.029, 0.172] \\
GPT-5.5 & Low & \textbf{0.152} & [0.088, 0.217] & \textbf{0.172} & [0.109, 0.242] \\
GPT-4.1 & Default & $-0.140$ & [$-0.222$, $-0.043$] & 0.053 & [$-0.030$, 0.124] \\
GPT-4o & Default & $-0.121$ & [$-0.197$, $-0.035$] & $-0.019$ & [$-0.087$, 0.060] \\
Gemini-2.5-Pro & Default & 0.005 & [$-0.057$, 0.080] & 0.074 & [0.014, 0.142] \\
Claude-3.7-Sonnet & Default & $-0.051$ & [$-0.122$, 0.017] & 0.004 & [$-0.068$, 0.073] \\
Qwen3-32B & Default & $-0.043$ & [$-0.116$, 0.031] & 0.025 & [$-0.051$, 0.099] \\
Llama-3.3-70B & Default & $-0.080$ & [$-0.151$, $-0.003$] & $-0.001$ & [$-0.071$, 0.070] \\
Qwen3-8B & Default & 0.041 & [$-0.026$, 0.110] & 0.024 & [$-0.048$, 0.092] \\
All-models & Default & $-0.062$ & [$-0.139$, 0.014] & 0.018 & [$-0.055$, 0.088] \\
\bottomrule
\end{tabular}}
\caption{Representative zero-shot direct-prediction results across proprietary and open-weight families. Within-CEFR $\rho$ uses only within-stratum ranks. CIs use 1,000 task-clustered bootstrap samples. GPT-5.5 Low is the descriptive best among the evaluated model-prompt configurations and is a prompt-sensitivity result, not a population-validated psychometric estimator or cross-validated model-selection estimate. Complete results are in Appendix~\ref{app:complete_results}.}
\label{tab:direct_population}
\end{table}

\begin{table*}[p]
    \centering
    \resizebox{\textwidth}{!}{
    \begin{tabular}{lccccc|ccccc}
    \toprule
    & \multicolumn{5}{c}{\textbf{Overall Spearman$\uparrow$}} & \multicolumn{5}{c}{\textbf{RMSE$\downarrow$}} \\
    \cmidrule(lr){2-6} \cmidrule(lr){7-11}
    \textbf{Model} & \textbf{Default} & \textbf{Low} & \textbf{Medium} & \textbf{High} & \textbf{Prompt ens.} & \textbf{Default} & \textbf{Low} & \textbf{Medium} & \textbf{High} & \textbf{Prompt ens.} \\
    \midrule
    Llama 2-7B & -0.013 & -0.020 & -0.019 & -0.028 & -0.037 & 0.587 & 0.408 & 0.422 & 0.449 & 0.289 \\
    Llama 2-13B & \phantom{-}0.043 & \phantom{-}0.038 & -0.084 & \phantom{-}0.010 & \phantom{-}0.031 & 0.378 & 0.413 & 0.370 & 0.376 & 0.278 \\
    Llama 3.1-8B & -0.146 & -0.120 & -0.123 & -0.142 & -0.177 & 0.288 & 0.331 & 0.257 & 0.339 & 0.251 \\
    Llama 3.2-1B & -0.087 & -0.009 & \phantom{-}0.064 & -0.013 & \phantom{-}0.025 & 0.574 & 0.726 & 0.635 & 0.663 & 0.493 \\
    Llama 3.2-3B & -0.004 & -0.017 & \phantom{-}0.064 & \phantom{-}0.042 & \phantom{-}0.051 & 0.416 & 0.467 & 0.417 & 0.430 & 0.259 \\
    Llama 3.3-70B & -0.080 & -0.123 & -0.110 & -0.103 & -0.163 & 0.221 & 0.234 & 0.229 & 0.287 & 0.229 \\
    Mistral-7B-v0.3 & -0.004 & \phantom{-}0.009 & \phantom{-}0.001 & -0.097 & -0.033 & 0.359 & 0.332 & 0.262 & 0.385 & 0.271 \\
    OLMo 2-7B & -0.069 & \phantom{-}0.010 & -0.111 & -0.121 & -0.092 & 0.279 & 0.446 & 0.325 & 0.369 & 0.236 \\
    OLMo 2-13B & \phantom{-}0.002 & \phantom{-}0.013 & -0.053 & -0.094 & -0.037 & 0.218 & 0.236 & 0.263 & 0.281 & 0.196 \\
    Phi-3-mini & \phantom{-}0.001 & \phantom{-}0.070 & \phantom{-}0.009 & \phantom{-}0.025 & \phantom{-}0.032 & 0.390 & 0.356 & 0.353 & 0.372 & 0.278 \\
    Phi-3.5-mini & \phantom{-}0.014 & -0.071 & -0.056 & -0.041 & -0.030 & 0.533 & 0.513 & 0.530 & 0.544 & 0.319 \\
    Phi-4-mini & -0.068 & \phantom{-}0.000 & -0.026 & -0.021 & -0.056 & 0.293 & 0.275 & 0.242 & 0.259 & 0.192 \\
    Phi-4 & -0.130 & -0.017 & -0.072 & \phantom{-}0.012 & -0.087 & 0.162 & 0.158 & 0.161 & 0.168 & 0.146 \\
    \midrule
    Qwen2.5-0.5B & \phantom{-}0.010 & \phantom{-}0.101 & \phantom{-}0.085 & -0.003 & \phantom{-}0.040 & 0.856 & 0.648 & 0.557 & 0.664 & 0.506 \\
    Qwen2.5-1.5B & \phantom{-}0.092 & -0.039 & \phantom{-}0.074 & \phantom{-}0.018 & \phantom{-}0.017 & 0.774 & 0.746 & 0.700 & 0.744 & 0.630 \\
    Qwen2.5-3B & -0.017 & -0.042 & \phantom{-}0.010 & -0.016 & -0.028 & 0.447 & 0.496 & 0.443 & 0.520 & 0.375 \\
    Qwen2.5-7B & -0.059 & -0.035 & -0.089 & -0.093 & -0.101 & 0.172 & 0.186 & 0.174 & 0.194 & 0.158 \\
    Qwen2.5-14B & -0.011 & -0.020 & -0.042 & -0.062 & -0.055 & 0.205 & 0.233 & 0.223 & 0.240 & 0.220 \\
    Qwen2.5-32B & -0.080 & -0.152 & -0.127 & -0.086 & -0.157 & 0.217 & 0.204 & 0.184 & 0.230 & 0.192 \\
    Qwen3-0.6B & \phantom{-}0.008 & -0.041 & \phantom{-}0.112 & \phantom{-}0.007 & \phantom{-}0.018 & 0.533 & 0.561 & 0.522 & 0.556 & 0.398 \\
    Qwen3-1.7B & \phantom{-}0.006 & -0.071 & -0.030 & \phantom{-}0.033 & -0.019 & 0.293 & 0.232 & 0.222 & 0.223 & 0.218 \\
    Qwen3-4B & -0.006 & -0.144 & -0.061 & -0.128 & -0.131 & 0.186 & 0.148 & 0.146 & 0.170 & 0.144 \\
    Qwen3-8B & \phantom{-}0.041 & \phantom{-}0.021 & \phantom{-}0.020 & \phantom{-}0.037 & \phantom{-}0.049 & 0.157 & 0.156 & 0.152 & 0.166 & 0.144 \\
    Qwen3-14B & -0.014 & -0.061 & \phantom{-}0.004 & -0.051 & -0.047 & 0.164 & 0.183 & 0.180 & 0.219 & 0.179 \\
    Qwen3-32B & -0.043 & -0.081 & \phantom{-}0.002 & -0.049 & -0.084 & 0.206 & 0.189 & 0.174 & 0.196 & 0.180 \\
    Qwen3-235B-A22B & -0.006 & \phantom{-}0.034 & -0.060 & -0.041 & -0.023 & 0.148 & 0.147 & 0.147 & 0.161 & 0.147 \\
    Qwen3.5-9B & \phantom{-}0.000 & \phantom{-}0.002 & -0.037 & -0.050 & -0.028 & 0.159 & 0.180 & 0.166 & 0.160 & 0.144 \\
    Qwen3.5-27B & -0.112 & \phantom{-}0.013 & \phantom{-}0.073 & -0.102 & -0.018 & 0.172 & 0.178 & 0.169 & 0.183 & 0.163 \\
    Qwen3.5-122B-A10B & -0.065 & \phantom{-}0.040 & \phantom{-}0.008 & -0.060 & -0.025 & 0.202 & 0.209 & 0.190 & 0.188 & 0.157 \\
    \midrule
    Claude 3.5 Haiku & -0.120 & -0.146 & -0.140 & -0.144 & -0.203 & 0.198 & 0.216 & 0.192 & 0.253 & 0.207 \\
    Claude 3.7 Sonnet & -0.051 & -0.128 & -0.012 & \phantom{-}0.007 & -0.098 & 0.188 & 0.218 & 0.187 & 0.201 & 0.195 \\
    Gemini 2.0 Flash & -0.072 & -0.019 & -0.073 & -0.143 & -0.113 & 0.175 & 0.156 & 0.150 & 0.175 & 0.157 \\
    Gemini 2.5 Flash & \phantom{-}0.006 & -0.025 & -0.063 & -0.067 & -0.062 & 0.236 & 0.190 & 0.176 & 0.266 & 0.206 \\
    Gemini 2.5 Pro & \phantom{-}0.005 & -0.044 & \phantom{-}0.013 & -0.015 & -0.005 & 0.172 & 0.160 & 0.164 & 0.189 & 0.164 \\
    GPT-3.5-Turbo & -0.021 & -0.033 & -0.046 & -0.071 & -0.082 & 0.274 & 0.310 & 0.266 & 0.284 & 0.248 \\
    GPT-4.1-mini & -0.028 & -0.034 & -0.013 & \phantom{-}0.000 & -0.049 & 0.175 & 0.184 & 0.181 & 0.182 & 0.174 \\
    GPT-4.1 & -0.140 & -0.171 & -0.131 & -0.160 & -0.193 & 0.193 & 0.189 & 0.172 & 0.199 & 0.182 \\
    GPT-4o-mini & -0.012 & -0.096 & -0.052 & \phantom{-}0.043 & -0.025 & 0.228 & 0.161 & 0.165 & 0.208 & 0.180 \\
    GPT-4o & -0.121 & -0.102 & -0.100 & -0.083 & -0.135 & 0.179 & 0.181 & 0.186 & 0.210 & 0.180 \\
    GPT-o4-mini & -0.101 & -0.161 & -0.153 & -0.074 & -0.184 & 0.146 & 0.150 & 0.143 & 0.146 & 0.141 \\
    GPT-5 & -0.071 & -0.051 & -0.131 & -0.082 & -0.127 & 0.160 & 0.160 & 0.165 & 0.163 & 0.157 \\
    GPT-5.5 & \phantom{-}0.129 & \phantom{-}0.152 & \phantom{-}0.120 & \phantom{-}0.071 & \phantom{-}0.163 & 0.153 & 0.153 & 0.151 & 0.157 & 0.151 \\
    \midrule
    All-models & -0.062 & -0.041 & -0.006 & -0.080 & -0.068 & 0.160 & 0.153 & 0.158 & 0.185 & 0.160 \\
    \bottomrule
    \end{tabular}
    }
\caption{Complete audited zero-shot direct-prediction results for all 42 models. Default denotes no persona. Low, Medium, and High denote prompt conditions. Prompt ens.\ averages item-level predictions across prompt conditions within each model, and All-models averages item-level predictions across models. Ensemble metrics are computed from these averaged item-level predictions rather than by averaging model- or prompt-level metrics. Values use the same scalar parser and validity rules as the main text. Population-aware results and task-clustered uncertainty are reported in Appendix~\ref{app:population_aware}.}
    \label{tab:table1_all}
\end{table*}

\begin{table}[p]
\centering
\small
\resizebox{\linewidth}{!}{
\begin{tabular}{lcccc}
\toprule
Method & Low@10 & Low@20 & High@10 & High@20 \\
\midrule
Llama 2-7B & 0.076 & 0.201 & 0.101 & 0.201 \\
Llama 2-13B & 0.063 & 0.208 & 0.177 & 0.201 \\
Llama 3.1-8B & 0.089 & 0.182 & 0.025 & 0.101 \\
Llama 3.2-1B & 0.089 & 0.176 & 0.076 & 0.195 \\
Llama 3.2-3B & 0.101 & 0.226 & 0.114 & \underline{0.270} \\
Llama 3.3-70B & 0.139 & 0.195 & 0.025 & 0.113 \\
Mistral-7B-v0.3 & 0.063 & 0.226 & 0.101 & 0.164 \\
OLMo 2-7B & 0.076 & 0.151 & 0.025 & 0.151 \\
OLMo 2-13B & 0.089 & 0.239 & 0.076 & 0.176 \\
Phi-3-mini & 0.076 & 0.176 & \underline{0.203} & 0.258 \\
Phi-3.5-mini & 0.127 & 0.195 & 0.063 & 0.138 \\
Phi-4-mini & 0.127 & 0.208 & 0.089 & 0.101 \\
Phi-4 & 0.076 & 0.170 & 0.025 & 0.101 \\
\midrule
Qwen2.5-0.5B & 0.013 & 0.138 & 0.089 & 0.239 \\
Qwen2.5-1.5B & 0.063 & 0.170 & 0.177 & 0.245 \\
Qwen2.5-3B & 0.127 & 0.201 & 0.114 & 0.164 \\
Qwen2.5-7B & 0.101 & 0.170 & 0.089 & 0.132 \\
Qwen2.5-14B & 0.063 & 0.195 & 0.089 & 0.132 \\
Qwen2.5-32B & 0.076 & 0.157 & 0.038 & 0.119 \\
Qwen3-0.6B & 0.051 & 0.176 & 0.139 & 0.214 \\
Qwen3-1.7B & 0.101 & 0.170 & 0.127 & 0.208 \\
Qwen3-4B & 0.101 & 0.182 & 0.025 & 0.094 \\
Qwen3-8B & 0.101 & 0.289 & 0.089 & 0.195 \\
Qwen3-14B & 0.089 & 0.252 & 0.025 & 0.151 \\
Qwen3-32B & 0.063 & 0.208 & 0.089 & 0.119 \\
Qwen3-235B-A22B & 0.101 & 0.182 & 0.051 & 0.182 \\
Qwen3.5-9B & 0.101 & 0.214 & 0.114 & 0.145 \\
Qwen3.5-27B & 0.139 & 0.252 & 0.038 & 0.157 \\
Qwen3.5-122B-A10B & 0.139 & 0.201 & 0.101 & 0.226 \\
\midrule
Claude 3.5 Haiku & 0.038 & 0.138 & 0.038 & 0.132 \\
Claude 3.7 Sonnet & 0.063 & 0.214 & 0.013 & 0.119 \\
Gemini 2.0 Flash & 0.038 & 0.189 & 0.051 & 0.151 \\
Gemini 2.5 Flash & 0.127 & 0.182 & 0.076 & 0.195 \\
Gemini 2.5 Pro & 0.177 & 0.233 & 0.038 & 0.113 \\
GPT-3.5-Turbo & 0.063 & 0.164 & 0.101 & 0.214 \\
GPT-4.1-mini & 0.101 & 0.182 & 0.076 & 0.182 \\
GPT-4.1 & 0.127 & 0.157 & 0.025 & 0.069 \\
GPT-4o-mini & 0.089 & 0.245 & 0.051 & 0.164 \\
GPT-4o & 0.089 & 0.157 & 0.051 & 0.101 \\
GPT-o4-mini & 0.101 & 0.189 & 0.025 & 0.119 \\
GPT-5 & 0.127 & 0.164 & 0.076 & 0.157 \\
GPT-5.5 & 0.190 & 0.296 & 0.139 & 0.252 \\
\midrule
All-models & 0.101 & 0.189 & 0.051 & 0.119 \\
\midrule
Item-total proxy, No-persona & \underline{0.342} & \textbf{0.377} & 0.165 & 0.252 \\
Item-total proxy, Low & 0.316 & 0.358 & 0.190 & \textbf{0.302} \\
Item-total proxy, Medium & 0.329 & 0.346 & 0.127 & 0.252 \\
Item-total proxy, High & 0.329 & \underline{0.371} & \textbf{0.278} & 0.264 \\
Item-total proxy, Full pool & \textbf{0.367} & \textbf{0.377} & \underline{0.203} & 0.277 \\
\midrule
Random & 0.100 & 0.200 & 0.100 & 0.200 \\
\bottomrule
\end{tabular}
}
\caption{Complete overlap-based retrieval results. Direct-prediction rows use LLM discrimination scores. Response-based rows use the audited uncorrected CEFR-stratified item-total proxy.}
\label{tab:table3_all}
\end{table}

\begin{table}[t]
\centering
\small
\resizebox{\linewidth}{!}{
\begin{tabular}{lrr}
\toprule
Corrected item-rest screening metric & Value & 95\% clustered CI \\
\midrule
Low@10 & 0.367 & [0.263, 0.468] \\
Low@20 & 0.377 & [0.321, 0.453] \\
Precision at 20\% review budget & 0.264 & [0.194, 0.344] \\
Recall of human discrimination $\leq0.2$ & 0.408 & [0.317, 0.500] \\
Enrichment over prevalence & $2.03\times$ & [$1.59\times$, $2.50\times$] \\
\bottomrule
\end{tabular}}
\caption{Task-clustered uncertainty for low-discrimination screening with the primary corrected full-pool item-rest estimator. The budget rows review the lowest-scored 20\% of items and define a target item by human discrimination $\leq0.2$.}
\label{tab:retrieval_uncertainty}
\end{table}

\end{document}